\pdfoutput=1
\documentclass[letterpaper]{article}
\usepackage{iftex}
\makeatletter
\providecommand{\NewCommandCopy}[2]{\let#1#2}
\@ifundefined{ifluatex}{\RequirePackage{ifluatex}}{}
\@ifundefined{iftutex}{%
  \newif\iftutex
  \ifLuaTeX\tutextrue\else\ifXeTeX\tutextrue\fi\fi
}{}
\makeatother
\makeatletter
\let\BJ@addcontentsline\addcontentsline
\makeatother
\usepackage[preprint]{aaai2027}
\usepackage[hyphens]{url}
\usepackage{graphicx}
\usepackage{natbib}
\usepackage{booktabs}
\usepackage{amsmath}
\usepackage{amssymb}
\usepackage{tikz}
\usetikzlibrary{arrows.meta,positioning,calc}
\usepackage{placeins}

\definecolor{mopgreen}{HTML}{12703F}
\definecolor{mopred}{HTML}{B32740}
\definecolor{mopblue}{HTML}{215FC4}
\definecolor{moppurple}{HTML}{5B4AB2}
\definecolor{mopmuted}{HTML}{55555F}
\definecolor{mopbg}{HTML}{F7F7F2}

\ifdefined\pdfsuppresswarningpagegroup
\fi
\title{Branch-JEPA: Finite-Support Predictive Distributions for JEPA World Models}
\author{
Zhi Song\textsuperscript{\rm 1,\rm 2},
Ximing Xing\textsuperscript{\rm 2},
Zhenchao Tang\textsuperscript{\rm 2},
Hanbo Huang\textsuperscript{\rm 2},
Jiehui Huang\textsuperscript{\rm 2},
Weilong Yan\textsuperscript{\rm 2},
Tianxu Lv\textsuperscript{\rm 2},
Minghao Yang\textsuperscript{\rm 2},
Zhongzheng Niu\textsuperscript{\rm 2},
Fan Xu\textsuperscript{\rm 2},
Bing He\corresponding\textsuperscript{\rm 2},
Lusheng Wang\corresponding\textsuperscript{\rm 1},
Jianhua Yao\corresponding\textsuperscript{\rm 2}
}
\affiliations{
\textsuperscript{\rm 1}Department of Computer Science, City University of Hong Kong, Hong Kong, China\\
\textsuperscript{\rm 2}Tencent, Shenzhen, 518057, China
}

\begin{document}
\maketitle

\begin{abstract}
Joint-embedding predictive architectures (JEPAs) learn dynamics by predicting future
observations in representation space.  Yet most JEPA world models return one latent
successor, even when hidden intent, partial observation, or stochastic dynamics make
several futures plausible.  We introduce Branch-JEPA, which replaces this point-valued
transition with a context-weighted finite set of latent successors.  Every branch is
decoded independently, and the complete set is retained at inference.  The architecture
supports two complementary training regimes: specialization for recovering separated
successors and full-set Energy-Score training for distributional fidelity.  In a locked
five-seed evaluation on the Argoverse~2 official validation split, full-set training
improves trajectory Energy Score by $5.8$--$6.5\%$ and probability-weighted trajectory
distance by $9.3$--$10.4\%$ over matched-$K{=}6$ assignment and transport objectives,
while retaining $5.36$ endpoint-deduplicated effective branches.  In a parameter-exact official-validation
comparison, latent branching retains $10.3\%$ more effective modes and improves Energy
Score, expected ADE, and Brier in all five paired seeds over branching only at the output
decoder; every paired 95\% interval excludes zero.  In an OGBench graph audit,
Branch-JEPA increases teleport verified-route existence to $19.2\%$ versus
$3.9\%$ for the MDN.  Its raw-support advantage also persists with 29-D state and RGB
observations.  Together, latent branching preserves more distinct futures, while
full-set scoring improves the quality of the resulting predictive distribution.
\end{abstract}

\section{Introduction}

World models should represent the futures that remain possible given an agent's current
information.  JEPAs learn such dynamics by predicting encoded targets rather than
reconstructing every observation
\citep{lecun2022path,assran2023ijepa,bardes2024vjepa}.  This abstraction has made them a
natural basis for latent world models \citep{zhou2024dinowm,assran2025vjepa2}.  Most JEPA
predictors, however, return a single latent future.  That is sufficient when the observed
history and action determine the next state, but not when hidden intent, partial
observation, or stochastic dynamics leave several futures plausible.

The limitation is structural.  A point prediction can summarize an ambiguous transition,
but it cannot enumerate separated outcomes.  Changing the point loss changes the
summary, not its cardinality; fusing the outputs of a gated mixture has the same
restriction.  We call this the \emph{point-valued support bottleneck}.  It concerns what
the predictor exposes, not whether its encoder has learned informative features.

Adding heads alone does not solve the problem.  Heads can duplicate one another, ignore
rare futures, or spread into unsupported regions.  A useful multi-future predictor must
both preserve distinct candidates and learn where their probability mass belongs.
Multiple-choice and multimodal prediction expose the resulting trade-off: hard
assignment encourages specialization but updates only the selected location, whereas
joint distributional training must balance coverage and fidelity
\citep{guzman2012multiple,lee2016stochastic,rupprecht2017mhp,%
makansi2019overcoming,narayanan2021dac,seo2020tmcl}.

Branch-JEPA makes multiplicity part of the JEPA transition.  One encoded context produces
$K$ latent successors and context-only weights.  The same decoder maps each successor
independently, preserving branch identity through deployment instead of averaging the
candidates back into one prediction.

How the branches are trained is a separate choice.  We study two regimes for the same
architecture.  \emph{Specialization} uses hard assignment to recover separated
successors.  \emph{Full-set} training scores every branch and its probability mass with
a proper multivariate score.

\begin{figure*}[t]
\centering
\includegraphics[width=0.98\textwidth]{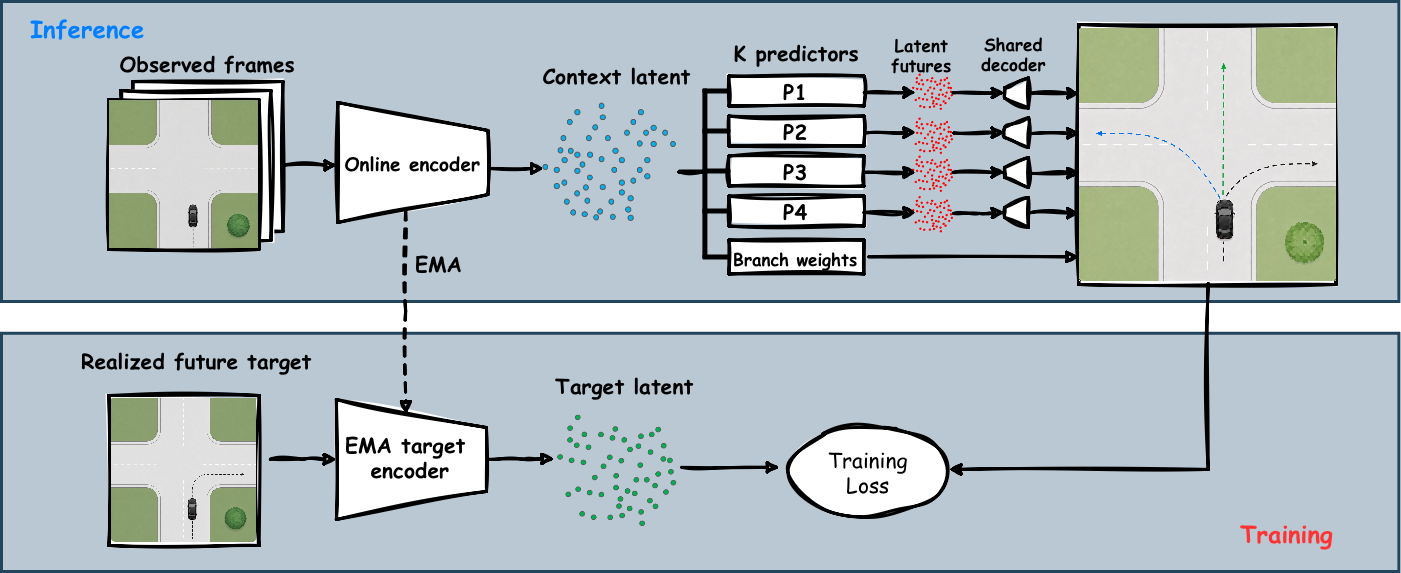}
\caption{\textbf{Branch-JEPA architecture.}
The deployable path encodes the observed context once, produces $K$ weighted latent
successors, and decodes each one independently with the same decoder; candidates are
never averaged.  During training, the EMA target encoder represents the realized future,
which supervises the branches through either specialization
(Eq.~\ref{eq:specialization}) or full-set scoring (Eq.~\ref{eq:av2}).
The target path is removed at inference, where both regimes return the complete weighted
set.  Four branches are drawn for clarity; road scenes and latent point clouds are
schematic.}
\label{fig:overview}
\end{figure*}

We test the architecture and training regimes separately.  An exactly parameter-matched
control keeps a single latent successor and branches only in the output decoder.
On official validation, Branch-JEPA retains $10.3\%$ more effective modes in this
comparison and improves Energy Score, expected ADE, and Brier in all five paired seeds.
Matched-$K$ experiments on
Argoverse~2 compare the two regimes' support--fidelity trade-offs, while OGBench
\citep{park2025ogbench}, AntMaze, and RGB diagnostics test whether specialization exposes
successor structure beyond motion forecasting.  We evaluate fidelity, calibration, effective support, and
map-aware quality rather than nominal branch count alone.
Figure~\ref{fig:overview} summarizes the architecture.

This study makes three contributions:
\begin{itemize}
\item We introduce a branch-preserving JEPA transition whose weighted latent successors
remain distinct through shared decoding; a parameter-exact control isolates its advantage
over branching only at the output decoder.
\item We formulate two complementary training regimes: hard specialization for separated
successors and latent/decoded full-set scoring for distributional fidelity, with a
population characterization as a $K$-atomic energy-discrepancy approximation.
\item We validate the architecture and objectives through locked matched-$K$ AV2
experiments, frozen-anchor and official-map audits, and specialization diagnostics on
OGBench, AntMaze, and RGB observations.
\end{itemize}
Throughout the paper, ``Branch-JEPA'' is the method name.  Full-set and specialization
name training objectives only, and are shown only where the objectives are compared.

\section{Related Work}

\paragraph{JEPA and predictive uncertainty.}
I-JEPA and V-JEPA established prediction in representation space
\citep{assran2023ijepa,bardes2024vjepa}; DINO-WM and V-JEPA~2 extend latent prediction
toward world modeling and planning
\citep{zhou2024dinowm,assran2025vjepa2,terver2026jepawms}.  Probabilistic formulations use
variational predictions \citep{huang2026varjepa,gogl2026varjepa}, while the concurrent
UWM-JEPA preprint carries a belief through blind rollout \citep{radha2026uwmjepa}.
D-JEPA and the concurrent JEDI preprint instead use generative prediction
\citep{chen2025djepa,lim2026jedi}.  Branch-JEPA returns an enumerable, fixed-budget atomic
set in one pass and scores that same measure on both sides of a shared decoder.  M3-JEPA
targets cross-modal alignment with multi-gate experts \citep{lei2025m3jepa}; its
weighted-sum readout motivates our fused-output control.

\paragraph{Multiple hypotheses and trajectory distributions.}
Multiple-choice and multiple-hypothesis learning specialize several outputs through hard
assignment \citep{guzman2012multiple,lee2016stochastic,rupprecht2017mhp}; trajectory-wise
MCL applies it to dynamics models \citep{seo2020tmcl}.  Later work documents collapsed or
spurious hypotheses and proposes annealing, data association, or diversity-aware sampling
\citep{makansi2019overcoming,narayanan2021dac,yuan2020dlow}.  Finite trajectory
distributions are represented by probabilistic anchors in MultiPath and discrete sets in
CoverNet \citep{chai2020multipath,phanminh2020covernet}, and by learned intention queries
in MTR \citep{shi2022mtr}.  ModeSeq sequentially decodes modes with target-matched
assignment \citep{zhou2025modeseq}.  MDNs \citep{bishop1994mixture}, Trajectron++
\citep{salzmann2020trajectron}, and MotionDiffuser \citep{jiang2023motiondiffuser} offer
continuous or sampled alternatives.  Forecast-MAE, QCNet, and Wayformer are stronger
forecasting backbones \citep{cheng2023forecast,zhou2023qcnet,nayakanti2023wayformer};
Branch-JEPA isolates the complementary question of how a JEPA predictor should represent
conditional support.  Scenario-aware prediction and contingency planning
\citep{cui2021lookout,chen2022scept,chen2023tpp} provide natural consumers of this
enumerable interface.  Hard assignment supplies Branch-JEPA's specialization regime; the
contribution is the branch-preserving JEPA architecture and its evaluation alongside
full-set distributional training, not a new assignment rule.

\paragraph{Full-distribution scoring.}
The Energy Score is a proper multivariate score \citep{gneiting2007strictly}.  DISCO Nets
learn sample-based distributions \citep{bouchacourt2016disco}; SampleNet learns a finite
empirical distribution with Energy-Score and Sinkhorn training
\citep{harakeh2023samplenet}.  Related uses include multivariate forecasting
\citep{kan2022mqf2}, trajectory evaluation \citep{shahroudi2024energy}, visual
autoregression \citep{shao2025continuous}, and diffusion transitions
\citep{debortoli2025distributional}.  Branch-JEPA differs by coupling one weighted atomic
measure across an EMA-target latent space and a shared decoded space, yielding both a
JEPA-native distributional target and a stable atom identity from latent prediction to
decoded deployment.  Our matched-$K$ comparison holds the Branch-JEPA architecture fixed
while comparing full-set scoring, uniform masses, hard and soft assignment, partial
optimal transport, and a development-tuned likelihood mixture.

\section{Branch-JEPA}

\paragraph{JEPA variables.}
For sample $i$, let $s_i$ be the observed context, $a_i$ an optional action, and
$s_i^+$ the future block.  The online encoder $f_\theta$ and its exponential-moving-average
target copy $f_\xi$ produce
\begin{equation}
\begin{aligned}
h_i&=f_\theta(s_i),&
z_i^+&=\operatorname{sg}(\operatorname{norm}(f_\xi(s_i^+))),\\
c_i&=[h_i,a_i].&&
\end{aligned}
\label{eq:jepa}
\end{equation}
where $\operatorname{sg}$ stops gradients and action-free domains set $c_i=h_i$.  After each
optimizer step, $\xi\leftarrow\tau\xi+(1-\tau)\theta$.  A standard predictor maps $c_i$ to
one normalized vector.  Its point-valued output cannot enumerate separated conditional
successors; averaging gated experts does not change that fact.

\paragraph{Point-valued support bottleneck.}
For a fixed context $c$, the population minimizer of squared point prediction is
\begin{equation}
g^*(c)=\arg\min_g\mathbb E[\lVert g-Z^+\rVert_2^2\mid c]
      =\mathbb E[Z^+\mid c].
\label{eq:conditional-mean}
\end{equation}
If predictions are constrained to the unit sphere and trained with cosine distance, the
minimizer is the normalized direction of this conditional mean whenever it is nonzero.
Equation~\ref{eq:conditional-mean} is the common squared/cosine special case, not a
characterization of every point loss.  Under a generic distance $D$, including the
$K{=}1$ Energy-Score control used below, the optimizer is instead a one-point Fr\'echet
representative
$g_D^*(c)\in\arg\min_g\mathbb E[D(g,Z^+)\mid c]$.
Mean or Fr\'echet representative, a singleton cannot enumerate separated conditional
support even when the encoder retains it.  A $K$-atomic output removes this cardinality
restriction because it can place different atoms in different regions.  This expressivity
statement does not imply that a particular loss or optimizer will discover every region;
effective support must be measured empirically.

\paragraph{Multi-branch predictor and shared inference contract.}
Branch-JEPA uses $K$ unfused predictors $g_{\phi_k}$ and a context-only router $r_\psi$:
\begin{equation}
u_{ik}=\operatorname{norm}(g_{\phi_k}(c_i)),\qquad
\pi_{ik}=\operatorname{softmax}(r_\psi(c_i))_k,
\label{eq:branches}
\end{equation}
which define
\begin{equation}
Q_{\theta,\phi,\psi}(\cdot\mid c_i)=\sum_{k=1}^{K}\pi_{ik}\delta_{u_{ik}}.
\label{eq:atomic}
\end{equation}
The router receives only $c_i$ and parameterizes context-dependent event weights.  At
inference, Branch-JEPA returns the full set
$\{(u_{ik},\pi_{ik})\}_{k=1}^K$: alternatives remain individually addressable rather than
being collapsed into a weighted average.

\paragraph{Decoded branches and block distance.}
Let $Y_i^+=(y_{i1}^+,\ldots,y_{iH}^+)$ be a future trajectory block.  Latent atom $u_{ik}$
denotes the entire block, and a shared decoder produces
$\widehat Y_{ik}=d_\omega(u_{ik})$.  For decoded trajectories and
$0<\beta\leq1$ we use
\begin{equation}
D_\beta(X,Y)=
\left(\frac{\sum_{t=1}^{H}\beta^{t-1}\lVert x_t-y_t\rVert_2^2}
{\sum_{t=1}^{H}\beta^{t-1}}\right)^{1/2};
\label{eq:block-distance}
\end{equation}
for normalized latents, $D_z(u,z)=\lVert u-z\rVert_2$.

\paragraph{Two training regimes.}
Branch-JEPA fixes the multi-branch architecture and its set-valued inference contract;
we study two ways to train it.  \emph{Specialization training} selects the closest branch
for each realized target,
\begin{equation}
\begin{aligned}
k_i^\star&=\arg\min_k D(X_{ik},T_i),\\
\mathcal L_{\mathrm{spec}}^{D}
&=\frac{1}{|\mathcal B|}\sum_{i\in\mathcal B}D(X_{i k_i^\star},T_i)
-\frac{\lambda_r}{|\mathcal B|}\sum_{i\in\mathcal B}\log\pi_{i k_i^\star}.
\end{aligned}
\label{eq:specialization}
\end{equation}
This classical hard-assignment rule drives explicit division of labor among branches; in
the AV2 instantiation, the same selected branch receives latent and decoded regression
updates.  \emph{Full-set training} instead treats the output as one weighted atomic
distribution.  Given support $X_{ik}$, target $T_i$, weights $\pi_{ik}$, and either
distance, write $d^T_{ik}=D(X_{ik},T_i)$ and
$d^X_{ikl}=D(X_{ik},X_{il})$.  Its empirical Energy Score is
\begin{equation}
\begin{aligned}
\mathcal L_{\mathrm{ES}}^{D}
={}&\frac{1}{|\mathcal B|}\sum_{i\in\mathcal B}
\bigg[\sum_{k=1}^{K}\pi_{ik}d^T_{ik}\\[-1mm]
&\qquad-\frac12\sum_{k,l=1}^{K}
\pi_{ik}\pi_{il}d^X_{ikl}\bigg].
\end{aligned}
\label{eq:energy}
\end{equation}
The observation term penalizes mass far from the realized target.  The pairwise term evaluates
the spread of the complete weighted support; duplicated atoms receive no pairwise credit.
Unlike ad hoc repulsion, the two terms form one proper distributional score.  Every
positive-mass atom appears in both terms, although atoms with near-zero mass and
nonconvex optimization can still yield effective under-utilization.  The two regimes
therefore share encoders, branches, router, decoder, and prediction-time output; only
their learning signal differs.

\paragraph{Population target.}
For a fixed context, let $P$ be the true conditional law and let
$\mathcal A_{\leq K}$ contain all probability measures with at most $K$ atoms.
On a compact strong-negative-type metric space $(\mathcal X,D)$
\citep{gneiting2007strictly,lyons2013distance}, define the
energy discrepancy
\[
\mathcal E_D(P,Q)=2\mathbb E D(X,Y)-\mathbb E D(X,X')
                 -\mathbb E D(Y,Y'),
\]
where $X,X'\stackrel{\mathrm{iid}}{\sim}Q$ and
$Y,Y'\stackrel{\mathrm{iid}}{\sim}P$.  Define
$\mathrm{ES}_D(Q,y)=\mathbb E D(X,y)-\frac12\mathbb E D(X,X')$.  Then
\begin{equation}
\mathbb E_{Y\sim P}\mathrm{ES}_D(Q,Y)
=\tfrac12\mathcal E_D(P,Q)+\tfrac12\mathbb E D(Y,Y').
\label{eq:energy-distance}
\end{equation}
Consequently, a population minimizer over $\mathcal A_{\leq K}$ exists and is
an energy-discrepancy-optimal weighted atomic approximation to $P$.  If
$P\in\mathcal A_{\leq K}$, the optimal measure is uniquely $P$, although its
slots remain non-identifiable under permutation or co-located mass splitting.
Outside that well-specified case, the projection need not be unique.  This
statement gives the unrestricted finite-support objective a population target;
it neither guarantees semantic mode recovery nor global optimization by a
shared neural parameterization.  Both $D_z$ and $D_\beta$ have strong negative
type because they are Euclidean metrics after restriction or a positive linear
rescaling.  The compactness premise is automatic for the unit-sphere latent
space.  In decoded Euclidean space, the identity and strict identification require
finite first $D_\beta$ moments; existence additionally requires a compact range or an
appropriate coercivity condition.  The proof is in the supplement.

The joint AV2 instantiation uses $\beta=1$ and
\begin{equation}
\begin{aligned}
\mathcal L_{\mathrm{AV2}}
={}&\lambda_z\mathcal L_{\mathrm{ES}}^{D_z}(u,z^+)
 +\lambda_y\mathcal L_{\mathrm{ES}}^{D_1}(\widehat Y,Y^+)\\
 &+\lambda_{\mathrm{rec}}\mathcal L_{\mathrm{rec}},
\end{aligned}
\label{eq:av2}
\end{equation}
where
$\mathcal L_{\mathrm{rec}}=\operatorname{SmoothL1}
(d_\omega(\operatorname{sg}(z^+)),Y^+)$ and
$\lambda_z=\lambda_y=\lambda_{\mathrm{rec}}=1$.
$D_z$ is a dimensionless latent chord distance, $D_1$ is a root-mean-square
trajectory distance in meters, and SmoothL1 is averaged over batch, time, and coordinate dimensions.
The fixed coefficients and normalization conventions are shared by every full-set arm.
The latent score updates the online encoder, predictor, and router.  The decoded score also
updates the shared decoder; reconstruction updates only that decoder.  The target encoder
moves only by EMA.  For $K=1$, the pairwise term vanishes, yielding a one-point Fr\'echet
predictor under the same objective rather than necessarily the arithmetic conditional mean.
The matched-$K$ study compares specialization and full-set training within Branch-JEPA;
the point and parameter-exact output-only controls test output cardinality and the
placement of multiplicity, respectively.

\section{Experiments}

\subsection{Experimental Questions and Protocol}

\paragraph{Evidence questions.}
We organize the evidence by question rather than by dataset chronology:
\textbf{(Q1)} what is gained by replacing a point transition with latent branches,
and do they add value beyond output-only branching; \textbf{(Q2)} how do specialization
and full-set training trade mode separation, fidelity, and effective support; and
\textbf{(Q3)} does broader candidate support persist across trajectory, state, and RGB
observations?  AV2 answers Q1--Q2 with matched controls; OGBench tests successor support
and transition fidelity, while AntMaze and RGB test raw candidate coverage.

\paragraph{AV2 evaluation protocol.}
All candidates and weights depend only on the observed scene.  Although specialization
uses the realized target to assign a training branch, neither regime uses that target to
select, remove, or alter candidates at inference.  We evaluate the complete
distribution with joint-trajectory and endpoint Energy Scores.  Fixed stop/straight/left/right
endpoint bins convert candidate mass into event probabilities, evaluated with Brier score
\citep{brier1950verification} and expected calibration error (ECE)
\citep{guo2017calibration}.  Stop means endpoint displacement below 2\,m; otherwise the
polar angle of that displacement defines left at $\geq30^\circ$, right at
$\leq-30^\circ$, and straight between those thresholds.
Brier is the per-scene sum over four classes, and top-label ECE uses 10 equal-width bins.
Router-top-1 ADE is mean Euclidean displacement over the 60 future steps, FDE is terminal
displacement, and miss rate is $\mathbf 1\{\mathrm{FDE}>2\,\mathrm m\}$.  They evaluate
$\arg\max_k\pi_k$ and therefore require no hindsight choice.  Full-distribution scores
evaluate the weighted conditional law, whereas router-top-1 and $K{=}1$ errors evaluate
a single committed prediction.

\paragraph{Effective support and score decomposition.}
Nominal $K$ is not evidence of multiplicity.  We therefore report entropy-effective
branch count, heads carrying at least $1\%$ aggregate mass, pairwise diversity, and a
collision-adjusted count.  The last is the inverse probability that two router samples
land within latent distance $\epsilon=.10$; it equals one for duplicate atoms and is not a
training objective.  We also decompose Energy Score into its expected observation distance
and dispersion credit.  Full definitions are in the supplement.
Minimum-of-$K$ errors are used only as AV2 coverage diagnostics, never for deployment-time
selection.

\begin{figure*}[!t]
\centering
\includegraphics[width=0.99\textwidth]{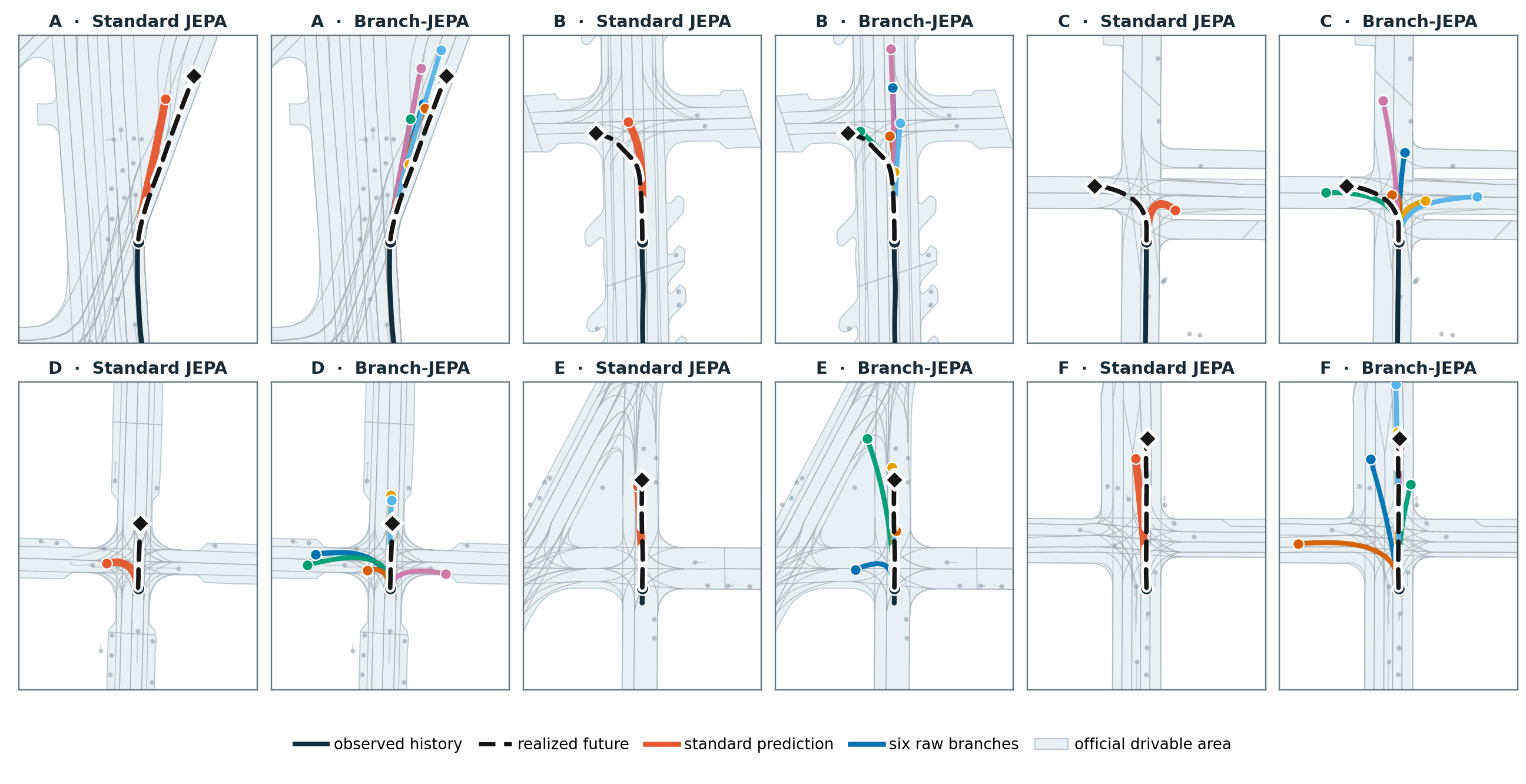}
\caption{\textbf{Branch-JEPA adapts its finite support to contextual ambiguity on AV2.}
Six target-blind cases from the frozen-anchor control are shown in a common agent-centric
frame.  Each adjacent pair contrasts Standard JEPA (orange, $K{=}1$) with all six raw
Branch-JEPA trajectories (colored).  The black dashed trajectory is the realized future,
added only after scenario identifiers were locked; it was unavailable to either predictor
and unused for case selection.  In (A), branches concentrate on one broad continuation;
in (B)--(F), they separate along distinct map-supported directions.  Every candidate
is shown; a separate target-blind official-map audit confirms that all six candidates
in every displayed case remain inside the official drivable polygons,
and no output is projected, snapped, filtered, or removed.  Population distribution and
containment results appear in Table~\ref{tab:joint} and the official-map audit.}
\label{fig:av2-supports}
\end{figure*}

\paragraph{Data and implementation.}
We use the official Argoverse~2 motion-forecasting data
\citep{wilson2021argoverse2}: 5\,s of focal-agent and neighbor history plus local map context
predict a 6\,s, 60-step future.  Deterministic preprocessing yields $199{,}908$ training and
$24{,}988$ validation scenes in a focal-centric frame aligned to the last observed heading.
We retain the 16 nearest observed neighbors and 48 nearest lane/crosswalk polylines, each
resampled to 10 points.  A one-layer GRU encodes the focal track; two-layer ReLU MLPs encode
flattened neighbor tracks and polylines; one four-head attention layer pools the 64 context
elements with the focal embedding as query.  A two-layer ReLU projection produces a
512-dimensional latent ($h=256$).  Each Branch-JEPA atom uses a 512--512--512 ReLU MLP;
the $K{=}1$ control uses 512--3204--512 to match predictor parameters.  The router is
512--256--$K$, and the shared decoder is 512--512--120.

The online view contains the 50 observed focal states, observed neighbors, and static map.
The target view replaces the focal track with the 60-step future; its velocity is the
finite difference divided by 0.1\,s and its heading is the velocity direction.  Future
neighbors are zeroed and masked apart from one zero dummy slot used only to avoid empty
attention, while the same static map is retained.  Online and target
encoders are separate instances of the identical architecture; the target is initialized
from the online encoder and updated only by EMA.  The GRU accepts the 50- and 60-step focal
sequences without an input-shape adapter.  Full feature definitions and preprocessing
details are in the supplement.

For the primary matched-$K$ study, we deterministically split the official training set
into 189,964 fitting and 9,944 development scenes; the 24,988-scene official validation
set is excluded from model, recipe, and checkpoint selection.  We train seeds
$\{0,\ldots,4\}$ for exactly 30 epochs without
early stopping, using AdamW, learning rate $3{\times}10^{-4}$, weight decay $10^{-2}$,
batch size 256 per GPU process, two linear-warmup epochs followed by cosine decay, mixed
precision, gradient norm 5, and EMA momentum $\tau=.996$.  Every arm uses 22,290
optimizer/EMA updates; within each seed, initialization, split, backbone, decoder, schedule,
and checkpoint selection by minimum development trajectory Energy Score are locked.
This protocol confines model and recipe selection to the train-derived development split.
After locking the complete arm--seed grid, we evaluate all 35 checkpoints once on the
official validation split without further tuning.  Only aggregate statistics are
retained; raw targets, raw predictions, and per-scene scores are not persisted.
The five parameter-exact output-only checkpoints use the same development-only
selection rule and are likewise evaluated once with the locked aggregate-only evaluator.
Two additive $K{=}6$ objective controls preserve every architectural and optimization
choice: soft-WTA uses one detached whole-trajectory responsibility vector at temperature
$.25$, while partial Sinkhorn mixes hard geometry with a balanced transport plan at
$\rho=.5$ and entropic regularization $\epsilon=.1$.  We also steelman the MDN over a
prespecified 12-configuration matrix spanning fixed, learned-global, and
component-specific scales, with and without component-usage regularization.  The
configuration with the lowest mean development trajectory Energy Score is selected
on development ($\sigma_z=.5,\sigma_Y=4$\,m); Table~\ref{tab:joint} reports its
locked official-validation result.
For finite-set evaluation, the MDN exposes its six component means and mixture masses;
Gaussian scales affect NLL training but are not sampled into extra candidates.
The earlier frozen-anchor attribution and map audit use seeds $\{0,1,2\}$ and the official
validation split for both selection and reporting as a separate secondary protocol; they
do not select any configuration in the primary locked evaluation.
The Standard JEPA baseline uses one predictor output ($K{=}1$) and is widened to
match the six-head predictor: total parameters differ by
130 ($0.0026\%$).  Counts include the online encoder, predictor/router, and decoder, but
exclude the training-only EMA target encoder.  The supplement documents the anchor gates and
evaluation definitions and per-seed results; the code archive contains preprocessing and
writes anchor hashes, resolved arguments, and training manifests during execution.

\subsection{Two Branch-JEPA Training Regimes at Matched Cardinality}

\begin{table*}[!t]
\centering
\small
\setlength{\tabcolsep}{1.5pt}
\begin{tabular}{@{}llrrrrrr@{}}
\toprule
model & objective & ES $\downarrow$ & obs. $\downarrow$ & exp.\ ADE $\downarrow$
& top-1 ADE $\downarrow$ & Brier $\downarrow$
& $N_{\rm eff}$ $\uparrow$\\
\midrule
\textbf{Branch-JEPA} & \textbf{full-set}
& $\mathbf{2.0766{\pm}.0057}$ & $\mathbf{3.6185{\pm}.0077}$
& $\mathbf{2.7941{\pm}.0059}$ & $\mathbf{2.4694{\pm}.0888}$
& $\mathbf{.09194{\pm}.00104}$ & $5.364{\pm}.023$\\
Branch-JEPA & full-set, uniform
& $2.1030{\pm}.0047$ & $3.6482{\pm}.0145$
& $2.8214{\pm}.0108$ & $2.7383{\pm}.4421$
& $.09316{\pm}.00099$ & $\mathbf{5.468{\pm}.050}$\\
Branch-JEPA & soft-WTA
& $2.2041{\pm}.0089$ & $3.9876{\pm}.0177$
& $3.0677{\pm}.0139$ & $2.4858{\pm}.0223$
& $.09352{\pm}.00054$ & $5.132{\pm}.023$\\
Branch-JEPA & Sinkhorn
& $2.2072{\pm}.0130$ & $3.9890{\pm}.0225$
& $3.0695{\pm}.0192$ & $2.5143{\pm}.0117$
& $.09388{\pm}.00047$ & $5.150{\pm}.038$\\
Branch-JEPA & specialization
& $2.2201{\pm}.0088$ & $4.0395{\pm}.0177$
& $3.1103{\pm}.0133$ & $2.4932{\pm}.0125$
& $.09360{\pm}.00057$ & $5.125{\pm}.009$\\
\midrule
MDN & NLL, tuned
& $2.3827{\pm}.0839$ & $4.0605{\pm}.1138$
& $3.1253{\pm}.0868$ & $2.5609{\pm}.0860$
& $.10130{\pm}.00318$ & $3.824{\pm}.111$\\
\bottomrule
\end{tabular}
\caption{\textbf{Matched-cardinality AV2 objective matrix}
(mean and sample SD across five locked seeds on the official validation split).
All $K{=}6$ rows share the EMA-target backbone, output count, initialization protocol,
training budget, and development-only checkpoint rule.  Observation is the
probability-weighted full-trajectory distance before the Energy Score dispersion credit;
router top-1 is target-blind.  Deduplication merges endpoint-connected components within
1\,m.  Full-set scoring gives the best mean Energy Score, observation
fidelity, expected ADE, router-top-1 ADE, and Brier; all six atoms remain active.
The MDN is the best mean development ES among a prespecified 12-configuration
scale matrix and is evaluated here without official-split retuning.
The parameter-matched $K{=}1$ structural control and the complete official-validation
per-seed matrix, including ECE and endpoint metrics, are reported in the supplement.}
\label{tab:joint}
\end{table*}

\paragraph{Full-set versus specialization.}
The first and fifth rows of Table~\ref{tab:joint} are two training regimes of the same
Branch-JEPA architecture.  Specialization keeps all six branches active and provides the
explicit mode-separation behavior used in our cross-domain experiments.  Full-set training
retains all six while lowering trajectory Energy Score by $6.46\%$, its observation
term by $10.42\%$, expected ADE by $10.16\%$, and Brier by $1.78\%$ relative to
specialization.  Against soft-WTA and partial Sinkhorn, its Energy Score improves by
$5.78/5.91\%$, its observation term by $9.26/9.29\%$, and expected ADE by
$8.92/8.97\%$.  Every Energy-Score and observation comparison has the same direction
in all five paired official-validation seeds.  Thus the gain is not carried only by
the dispersion credit: probability-weighted candidate-to-observation fidelity improves
at the same time.
For the direct full-set-minus-specialization comparison, the paired mean differences are
$-.1435$ trajectory ES (paired-$t$ 95\% interval $[-.1574,-.1295]$) and
$-.4211$ observation distance ($[-.4512,-.3909]$).  Full-set training also has the
best mean target-blind router-top-1 ADE among the matched-$K$ rows.

The 12-configuration MDN sweep narrows the earlier baseline gap but does not reverse the
ordering on locked official validation: full-set Branch-JEPA improves Energy Score,
the observation term, expected ADE, and Brier by $12.8\%$, $10.9\%$, $10.6\%$,
and $9.2\%$, respectively.  It also retains $5.36$ deduplicated effective
branches versus $3.82$ for the MDN.  These official-validation results preserve the
development ordering without official-split retuning.

Fixed uniform masses remain a strong control at $2.1030$.  Learned context weights add
a further $1.25\%$ trajectory-ES improvement, $0.82\%$ observation improvement, and
$1.31\%$ Brier improvement, with the ES and observation direction preserved in every
paired seed.  The dominant gain therefore comes from learning the complete support,
while context-dependent mass estimation adds a smaller, repeatable improvement.
Supplementary Table S3 reports the parameter-matched $K{=}1$ structural reference,
the complete official-validation per-seed matrix, calibration and point-error metrics,
and paired statistics; Supplementary Table S10 gives the development MDN scale matrix.

\subsection{Latent Branching versus Output-Only Branching}
We also fix output cardinality, downstream trainable parameters
($3{,}608{,}958$), five seeds, 30 epochs, and the trajectory Energy Score, but
replace the $K$ latent atoms by one predicted latent followed by a shared decoder trunk
and $K$ trajectory residual heads.  Table~\ref{tab:architecture-attribution} shows the
locked official comparison.  Latent branching improves Energy Score, expected ADE,
Brier, and effective support in every paired seed.  All four paired 95\% intervals
exclude zero; for Energy Score and effective modes the Branch-minus-output-only
intervals are $[-.0447,-.0304]$ and $[.456,.631]$.  Thus, at exactly matched
downstream capacity and output count, placing multiplicity in the JEPA transition
provides more effective support and better distribution quality than introducing it
only after a singleton latent.  Per-seed results and the exact parameter identity are
in Supplementary Table S4.

\begin{table*}[t]
\centering
\small
\setlength{\tabcolsep}{8pt}
\begin{tabular}{@{}lcrrr@{}}
\toprule
architecture & latent$\to$out & ES $\downarrow$
& Brier $\downarrow$ & $N_{\rm eff}$ $\uparrow$\\
\midrule
Standard JEPA & $1\to1$ & $2.8426{\pm}.0046$
& $.12095{\pm}.00160$ & $1.000$\\
output-only branching & $1\to6$ & $2.1142{\pm}.0046$
& $.09554{\pm}.00043$ & $5.254{\pm}.051$\\
\textbf{Branch-JEPA} & $\mathbf{6\to6}$ & $\mathbf{2.0766{\pm}.0057}$
& $\mathbf{.09194{\pm}.00104}$ & $\mathbf{5.797{\pm}.036}$\\
\bottomrule
\end{tabular}
\caption{\textbf{Official-validation architecture ladder} (five locked seeds).
The widened $K{=}1$ row is a parameter-matched structural reference.  The two
$K{=}6$ rows have exactly $3{,}608{,}958$ downstream trainable parameters and isolate
whether multiplicity is placed in the latent transition or only after one latent.
$N_{\rm eff}=\exp H(\pi)$.  In the locked same-device batch-64 audit, end-to-end
median latency is $2.89{\pm}.59$ versus $2.41{\pm}.22$\,ms per batch for
Branch-JEPA and $K{=}1$, respectively.}
\label{tab:architecture-attribution}
\end{table*}

\paragraph{Latent-objective ablation.}
On the train-derived development split, a five-seed same-architecture control removes
all target-path gradients and trains only
the decoded set.  Relative to this control, the full objective reduces expected ADE
from $2.7904{\pm}.0128$ to $2.7644{\pm}.0133$ and raises effective modes from
$5.593{\pm}.042$ to $5.806{\pm}.010$; the paired 95\% differences are
$[-.0476,-.0044]$ and $[.155,.271]$.  Its trajectory Energy Score is comparable
($2.0560$ versus $2.0577$).  Decoded full-set scoring therefore supplies the main
task-space score, while latent supervision improves candidate fidelity and support
utilization.

\subsection{Predictor-Only Frozen-Anchor Control}

To isolate the predictor from decoder adaptation, each seed first trains a future-view
target encoder and decoder; single-branch and full-set Branch-JEPA predictors then share
this identical SHA-locked, frozen anchor and receive no task-space gradient.
All three anchors pass prespecified variance and effective-rank gates
($r_{\rm ent}=12.86$--$13.20$), ruling out the measured target-collapse modes.
Full-set
Branch-JEPA improves latent, trajectory, and endpoint Energy Scores by $26.7\%$, $26.3\%$,
and $26.5\%$ over the parameter-matched single-branch predictor, while all six atoms
remain active.  Crucially,
collision-adjusted latent effective support is $4.26{\pm}.05$, versus 1 for the point
predictor, so the gain is not explained by duplicated heads.  Supplementary
Tables S6--S7 report every seed and the prespecified anchor acceptance gates.

\subsection{Official-Map Containment Audit}

In the locked primary evaluation, full-set Branch-JEPA places
$99.319{\pm}.041\%$ probability-weighted trajectory points inside official drivable
polygons, while retaining $5.399{\pm}.025$ mass-weighted map-feasible effective
branches.  The corresponding $95\%$-point-contained trajectory mass is
$97.157{\pm}.181\%$.
The frozen-anchor audit exports every candidate and router probability without loading the
realized future and restores outputs to world coordinates without projection or map
matching.  On $22{,}979$ AV2 validation scenes with a vehicle or bus focal actor,
full-set Branch-JEPA places $96.775{\pm}.153\%$ probability mass on
drivable-polygon-contained trajectories and retains $5.093{\pm}.027$ mass-weighted
contained effective support.  Its contained endpoint diversity is
$6.426{\pm}.065$\,m.  Thus the multi-atom output remains largely inside official
drivable polygons while retaining roughly five distinct contained alternatives.  The
full seven-metric audit and paired-scene bootstrap intervals are in the supplement; this
population audit complements the target-blind qualitative cases in
Figure~\ref{fig:av2-supports}.

\subsection{Specialization Results Across Domains}

\begin{table}[t]
\centering
\small
\setlength{\tabcolsep}{2.4pt}
\begin{tabular}{@{}lrrr@{}}
\toprule
metric (\%) & Standard & MDN & \textbf{Branch}\\
\midrule
verified route: medium & $8.3$ & $13.4$ & $\mathbf{24.7}$\\
verified route: teleport & $3.5$ & $3.9$ & $\mathbf{19.2}$\\
graph precision range & $62$--$90$ & $14$--$21$ & $49$--$56$\\
\midrule
29-D state raw COV-R & $29.4$ & $38.6$ & $\mathbf{57.6}$\\
RGB raw COV-R & $24.3$ & $26.7$ & $\mathbf{61.3}$\\
\bottomrule
\end{tabular}
\caption{\textbf{OGBench support and transition fidelity.}
The upper block uses the state-only $K{=}8$ graph protocol; the lower block reports
separate higher-dimensional support campaigns.  Entries are seed means; protocols,
SDs, and controls are in the supplement.  These are offline diagnostics, not closed-loop
control metrics.}
\label{tab:ogbench-interface}
\end{table}

Branch-JEPA trained with the mode-specializing objective raises macro transition
precision from $14$--$21\%$ to $49$--$56\%$, and verified-route existence improves
to $24.7\%$ versus $13.4\%$ on medium and $19.2\%$ versus $3.9\%$ on teleport.
The raw-coverage advantage also
persists for 29-D AntMaze state ($57.6\%$ versus $38.6\%$) and RGB observations
($61.3\%$ versus $26.7\%$); these latter results test breadth of candidate support, not
semantic modes.  Additional statistics and visualizations are in the supplement.

\section{Interpretation}

The evidence separates architecture from training regime.  Specialization exposes broad,
high-precision successor support on OGBench, whereas full-set training improves AV2
distribution quality and observation fidelity over hard/soft assignment and partial
transport while retaining all six branches.  The parameter-exact output-only control
shows on official validation a pronounced effective-support advantage and systematic
distribution-quality gains from placing multiplicity in the latent transition.
The frozen-anchor and map audits further show that this multiplicity belongs to the
predictor and survives shared decoding as roughly five distinct drivable-area-contained
futures.  Fixed-uniform weighting coming within $1.25\%$ indicates that branch locations
provide most of the gain, with context-dependent masses adding a smaller improvement.
The cross-domain results use specialization and remain support diagnostics, not evidence
for full-set training or closed-loop control.

\section{Scope}

Branch-JEPA targets the predictive-distribution interface rather than forecasting SOTA.
The primary AV2 configurations are selected on development and evaluated once on official
validation.  Map containment does not test lane connectivity, dynamics, or collision, and
the OGBench graph/set diagnostics are not closed-loop control.

\section{Conclusion}

Branch-JEPA replaces a point-valued JEPA transition with $K$ unfused successors that
survive shared decoding.  On official AV2 validation, full-set scoring improves
distribution quality over matched objectives.  A parameter-exact five-seed control
further shows that latent branching retains $10.3\%$ more effective modes and improves
Energy Score, expected ADE, and Brier over output-only branching.  Specialization
broadens successor support in OGBench, AntMaze, and RGB observations.

\begingroup
% Keep the submitted main-paper references immediately after its conclusion.
\bibliography{refs}
\endgroup

\clearpage
% Reset numbering and float-placement heuristics for the appended supplement.
\setcounter{secnumdepth}{1}
\setcounter{section}{0}
\setcounter{subsection}{0}
\setcounter{table}{0}
\setcounter{figure}{0}
\setcounter{equation}{0}
\renewcommand{\thesection}{S\arabic{section}}
\renewcommand{\thetable}{S\arabic{table}}
\renewcommand{\thefigure}{S\arabic{figure}}
\renewcommand{\theequation}{S\arabic{equation}}
% Keep the compact, full-width audit tables together instead of deferring them
% onto sparsely populated float pages.  A float page must be nearly full before
% LaTeX is allowed to build one, so short floats stay on text pages instead.
\setcounter{dbltopnumber}{4}
\setcounter{topnumber}{3}
\setcounter{totalnumber}{5}
\renewcommand{\dbltopfraction}{0.95}
\renewcommand{\topfraction}{0.95}
\renewcommand{\bottomfraction}{0.5}
\renewcommand{\floatpagefraction}{0.90}
\renewcommand{\dblfloatpagefraction}{0.90}
\renewcommand{\textfraction}{0.05}
\raggedbottom
% Restore the real \addcontentsline now, after every main-paper \section has
% already run, so only the supplement's headings populate the table of contents.
\makeatletter
\let\addcontentsline\BJ@addcontentsline
\makeatother
\twocolumn[
\begin{center}
{\LARGE\bfseries Supplementary Material\par}
\vspace{0.35em}
{\large Branch-JEPA: Finite-Support Predictive Distributions for JEPA World Models\par}
\end{center}
\vspace{0.8em}
]

\setcounter{tocdepth}{2}
% The default \numberline box is sized for a bare 1-2 digit section number;
% widen it so the "S10"/"S11" prefix does not collide with the entry title.
\makeatletter
\renewcommand{\l@section}{\@dottedtocline{1}{1.5em}{3.0em}}
\renewcommand{\l@subsection}{\@dottedtocline{2}{3.8em}{3.6em}}
\makeatother
\tableofcontents

\section{Submission Scope}

This supplement documents one Branch-JEPA architecture under its two training regimes.
AV2 compares specialization and full-set training at matched cardinality, while OGBench
and AntMaze use specialization training to test explicit support recovery across
representations.  These experiments answer complementary architecture- and
training-level questions: the cross-domain results do not compare the two objectives,
and the AV2 objective comparison does not establish cross-domain planning performance.
Historical ETH/UCY, SVHN, and Forecast-MAE exploratory
artifacts are excluded from the formal evidence package.

\section{AV2 Data Construction}

\subsection{Observed context}

We use the official Argoverse~2 motion-forecasting training and validation sets cited in the
main paper.  The cache contains $199{,}908$ training and $24{,}988$
validation scenes.  The focal coordinate frame has its origin at the focal agent's last
observed position (time index 49) and is rotated by the negative of its last observed heading.
All trajectories and map coordinates remain in meters.

The focal track contains 50 steps with features
\[
[x,y,v_x,v_y,\cos h,\sin h,\mathrm{type}],
\]
where the type flag is zero.  We retain the 16 closest tracks present at time index 49,
using the same 50-step layout and a binary non-vehicle flag.  Missing timesteps and unused
neighbor slots are zero-padded and masked.

For every lane segment, we average the left and right boundaries after resampling each to
10 points.  Pedestrian crossings use one resampled edge.  Each polyline point contains
\[
[x,y,\cos d,\sin d,\mathrm{intersection},\mathrm{crosswalk}],
\]
where $(\cos d,\sin d)$ is the local tangent.  We retain the 48 polylines with minimum
distance to the focal origin and mask unused slots.  Selection uses only the observed scene
and static map.

\subsection{Future target view}

The prediction target is the focal 60-step future trajectory $Y^+\in\mathbb R^{60\times2}$.
For the JEPA target encoder, we convert it to a 60-step focal feature sequence with the same
seven channels as the observed focal track.  Velocity is the forward position difference
divided by the AV2 interval $0.1$\,s; the last velocity repeats the preceding value.  Heading
cosine and sine are the normalized velocity direction, and the type flag is zero.  Future
neighbors are unavailable and are therefore all zero with a zero validity mask.  The static
map tensor and mask are copied from the observed context.  An all-empty-attention guard
leaves one zero-valued dummy neighbor slot unmasked; this contains no future information.

The online and target encoders are distinct instances of the same network.  A one-layer GRU
accepts the 50-step online focal view or 60-step target focal view directly; no learned
length-specific adapter is used.  The target output is L2-normalized and stop-gradient.  In
the joint model the target parameters are initialized from the online encoder and updated
after every optimizer step with EMA momentum $.996$.  The predictor and router never receive
the future view at inference.  The online context embedding itself is not L2-normalized
before the predictor or router; normalization is applied to the target latent and to each
predicted atom.

\section{Architecture and Optimization}

\begin{table}[t]
\centering
\small
\setlength{\tabcolsep}{4pt}
\begin{tabular}{@{}lll@{}}
\toprule
Module & Input/output & Exact construction \\
\midrule
Focal encoder & $7\!\to\!256$ & one-layer GRU \\
Neighbor encoder & $350\!\to\!256$ & Linear--512--ReLU--Linear \\
Polyline encoder & $60\!\to\!256$ & Linear--512--ReLU--Linear \\
Context pooling & $256\!\to\!256$ & one 4-head attention layer \\
Latent projection & $512\!\to\!512$ & Linear--512--ReLU--Linear \\
Each atom predictor & $512\!\to\!512$ & Linear--512--ReLU--Linear \\
Router & $512\!\to\!K$ & Linear--256--ReLU--Linear \\
Shared decoder & $512\!\to\!120$ & Linear--512--ReLU--Linear \\
\bottomrule
\end{tabular}
\caption{Exact architecture used by the joint and frozen-anchor AV2 experiments.  The
attention query is the focal GRU state; keys and values are the 16 neighbor plus 48
polyline embeddings.}
\label{tab:architecture}
\end{table}

The parameter-matched Standard JEPA uses a 512--3204--512 ReLU MLP, widening its single
predictor to the parameter budget of the six-atom predictor and router.  In the joint
experiment the Standard-JEPA and Branch-JEPA models
contain $5{,}074{,}428$ and $5{,}074{,}558$ trainable parameters.  In frozen-anchor Stage B
they contain $3{,}284{,}612$ and $3{,}284{,}742$ trainable parameters; the shared target
encoder and decoder are frozen and excluded from those counts.  Joint counts likewise
exclude the training-only EMA target encoder.

The output-only branching control keeps one 512--512--512 latent predictor and moves
cardinality to the task-space readout.  A context-only router predicts six masses; a
shared 512--2180 decoder trunk produces a base trajectory, and six residual heads produce
the six complete 60-step trajectories.  Reconstruction of the EMA target latent updates
only the shared base path, whereas the complete trajectory Energy Score updates the
trunk, residual heads, router, online encoder, and singleton latent predictor.  The
Branch-JEPA predictor/router/shared-decoder and this output-only
predictor/router/base-residual decoder each contain exactly $3{,}608{,}958$ downstream
trainable parameters (difference zero).  All 396 functional residual-scale parameters
receive nonzero gradients; no inactive padding parameters are used for matching.  Both
models emit six target-blind trajectories and use the same split, seeds, 30-epoch budget,
optimizer schedule, and development-Energy checkpoint rule.

All six primary $K{=}6$ rows emit six latent candidates through a shared backbone.
Branch-JEPA with learned-mass full-set training uses the two Energy Scores in the main
paper.  Its uniform-mass variant fixes every mass to $1/6$ and gives the router zero
gradient and no prediction-time role.  Branch-JEPA with specialization training selects
one winner by complete 60-step trajectory distance, applies latent and trajectory
regression to that same branch, and trains the router with its assignment label.  Both
Branch-JEPA regimes otherwise use the same multi-branch architecture and complete-set
inference.  The soft-WTA control computes one detached responsibility vector
$\operatorname{softmax}(-D_1/\tau)$ from each complete trajectory block at
$\tau=.25$ and reuses it for latent geometry, decoded geometry, and router supervision.
The partial-Sinkhorn control mixes the hard geometry plan with a balanced entropic
transport plan at $\rho=.5$ and $\epsilon=.1$; its router retains the hard
whole-trajectory label.  Both controls are target blind at inference.

The MDN arm minimizes mixture NLL in latent and trajectory space.  A prespecified
12-configuration matrix (Table~\ref{tab:mdn-steelman-grid}) contains nine fixed scale pairs
$\sigma_z\in\{.1,.25,.5\}$ and $\sigma_Y\in\{1,2,4\}$\,m, learned-global scales,
component-specific diagonal scales, and component-specific scales with a $.01$
usage-KL term.  The main table reports the lowest mean development Energy Score,
the fixed pair $\sigma_z=.5,\sigma_Y=4$\,m.  This is a baseline-favorable
development selection rather than an independent test estimate.  For common evaluation
we read out all six component means and predicted mixture masses without density sampling.

\begin{table}[!t]
\centering
\scriptsize
\setlength{\tabcolsep}{2.5pt}
\resizebox{\columnwidth}{!}{%
\begin{tabular}{@{}lrrrrrrr@{}}
\toprule
MDN scale family & $\sigma_z$ & $\sigma_Y$ (m) & usage KL
& ES-traj. $\downarrow$ & Brier $\downarrow$ & ECE $\downarrow$
& $\exp H(\pi)$ $\uparrow$\\
\midrule
component diagonal & .25 & 2 & 0 & 3.2942 & .19082 & .08763 & 5.10\\
component diagonal & .25 & 2 & .01 & 3.2703 & .18434 & .08218 & 5.14\\
\midrule
fixed isotropic & .10 & 1 & 0 & 3.5584 & .15998 & .07446 & 3.80\\
fixed isotropic & .10 & 2 & 0 & 3.2013 & .14046 & .06303 & 3.62\\
fixed isotropic & .10 & 4 & 0 & 2.8083 & .11179 & .03440 & 3.27\\
fixed isotropic & .25 & 1 & 0 & 3.2290 & .15509 & .06246 & 3.55\\
fixed isotropic & .25 & 2 & 0 & 3.2371 & .15633 & .07227 & 3.51\\
fixed isotropic & .25 & 4 & 0 & 2.7683 & .12620 & .05739 & 3.03\\
fixed isotropic & .50 & 1 & 0 & 3.1703 & .16809 & .05855 & 4.11\\
fixed isotropic & .50 & 2 & 0 & 3.1849 & .14886 & .05951 & 3.77\\
fixed isotropic & .50 & 4 & 0 & \textbf{2.3710} & \textbf{.10567} & .02854 & 3.84\\
learned global & .25 init. & 2 init. & 0 & 3.4750 & .12854 & \textbf{.02789} & 4.30\\
\bottomrule
\end{tabular}}
\caption{\textbf{Complete five-seed MDN steelman matrix.}
Entries are means on the same train-derived development split.  The fixed
$(\sigma_z,\sigma_Y)=(.5,4\,\mathrm m)$ row has the lowest mean trajectory Energy Score
and is the MDN row used in the main paper.  Because the row is selected on this development
matrix, it is deliberately favorable to the baseline and is not an independent test
estimate.}
\label{tab:mdn-steelman-grid}
\end{table}

The primary same-$K$ study deterministically partitions the 199,908 official training
scenes into 189,964 fitting and 9,944 development scenes.  For seeds
$\{0,\ldots,4\}$, every $K{=}6$ arm shares split membership, initialization seed, data
order, architecture, and schedule.  Each arm uses AdamW with learning rate
$3{\times}10^{-4}$ and weight decay $10^{-2}$, batch size 256 on one GPU, mixed precision,
gradient-norm clipping at 5, two linear-warmup epochs followed by cosine decay, and exactly
30 epochs.  Every run completes 22,290 optimizer steps and one EMA update per step; there
is no early stopping.  Among the 30 checkpoints, minimum train-derived-development
trajectory Energy Score selects one checkpoint.  The official validation set is not used
by these jobs for training, recipe selection, or checkpoint selection.
Selected epochs for seeds 0--4 are 29/30/28/29/30 (learned mass),
29/28/30/30/27 (uniform mass), 29/29/30/30/30 (hard assignment), and
29/28/30/30/30 (soft-WTA), 29/30/30/30/30 (partial Sinkhorn), and
29/29/30/29/28 (development-tuned MDN).
After this grid was locked, all 30 $K{=}6$ and five $K{=}1$ checkpoints were evaluated
in one pass over the 24,988 official-validation scenes.  The evaluator retained exactly
three aggregate JSON artifacts---a run summary, the five-seed table, and a completion
manifest---and persisted no raw validation target, raw prediction, or per-scene metric.
All aggregate files are SHA-256 bound to the fixed checkpoint grid and evaluator source.

The older frozen-anchor study uses seeds $\{0,1,2\}$, batch size 512, at most 30 epochs,
and patience 8.  It uses the official validation split for both checkpoint selection and
reporting and is therefore retained as exploratory secondary evidence rather than part of
the same-$K$ development protocol.
The original same-$K$ jobs use one internal cluster host with four NVIDIA H20 GPUs, with one
training process bound to each GPU.  The additive objective controls and MDN scale matrix use
packed eight-V100 jobs with one process per GPU and the same optimizer-step budget.  Every
job runs in one frozen container image providing the
Python~3.11/CUDA~12.4 runtime; the resolved training configuration is included in the code
supplement.

\begin{table*}[t]
\centering
\footnotesize
\setlength{\tabcolsep}{4.5pt}
\begin{tabular}{@{}p{.10\textwidth}p{.17\textwidth}p{.21\textwidth}p{.24\textwidth}p{.15\textwidth}@{}}
\toprule
Protocol & Target representation & Trainable modules & Training loss & Validation selection \\
\midrule
Joint / full-set & stop-grad EMA encoder
& online encoder, predictor/router, decoder
& latent ES $+$ trajectory ES $+$ target reconstruction
& trajectory ES \\
Joint / specialization & stop-grad EMA encoder
& online encoder, predictor/router, decoder
& assigned-branch latent/trajectory regression $+$ router CE $+$ target reconstruction
& trajectory ES \\
Anchor Stage A & directly trained future-view encoder
& future-view encoder, decoder
& reconstruction $+$ variance floor $+$ rank floor
& decoder ADE among gate-passing epochs \\
Anchor Stage B & within-seed frozen Stage-A encoder
& context encoder, predictor/router
& latent ES only
& decoded trajectory ES \\
\bottomrule
\end{tabular}
\caption{Gradient scopes.  Decoded trajectories in frozen-anchor Stage B are validation-only
and provide no training gradient.  Within a seed, the single-branch and full-set
Branch-JEPA arms load the same SHA-locked Stage-A checkpoint.}
\label{tab:gradient-scope}
\end{table*}

\section{Population Target of Full-Set Training}

We give the complete statement behind the full-set score in the main paper.  Let
$(\mathcal X,D)$ be a compact strong-negative-type metric space, and let $P,Q$ be Borel
probability measures on $\mathcal X$.  For
$X,X'\stackrel{\mathrm{iid}}{\sim}Q$ and
$Y,Y'\stackrel{\mathrm{iid}}{\sim}P$, define
\[
\begin{aligned}
\mathrm{ES}_D(Q,y)
&=\mathbb E D(X,y)-\tfrac12\mathbb E D(X,X'),\\
\mathcal E_D(P,Q)
&=2\mathbb E D(X,Y)-\mathbb E D(X,X')-\mathbb E D(Y,Y').
\end{aligned}
\]
Here $\mathcal E_D$ denotes the non-square-rooted energy discrepancy.  Taking expectation
of the first display over $Y\sim P$ and adding and subtracting
$\frac12\mathbb E D(Y,Y')$ gives
\[
\mathbb E_{Y\sim P}\mathrm{ES}_D(Q,Y)
=\tfrac12\mathcal E_D(P,Q)+\tfrac12\mathbb E D(Y,Y').
\]
The second term is constant in $Q$.  Strong negative type gives
$\mathcal E_D(P,Q)\geq0$, with equality if and only if $P=Q$.

Let
\[
\mathcal A_{\leq K}=
\left\{\sum_{k=1}^{K}\pi_k\delta_{x_k}:
(\pi_1,\ldots,\pi_K)\in\Delta_K,\ x_k\in\mathcal X\right\},
\]
where zero weights and co-located atoms allow support smaller than $K$.  At the measure
level, this class is the continuous image of the compact set
$\Delta_K\times\mathcal X^K$.  Because $D$ is continuous and bounded on the compact space,
$Q\mapsto\mathcal E_D(P,Q)$ is continuous, so a minimizer over
$\mathcal A_{\leq K}$ exists.  The identity above shows that its minimizers are exactly
the energy-discrepancy-optimal weighted atomic approximations to $P$.  If
$P\in\mathcal A_{\leq K}$, the minimizing probability measure is uniquely $P$; slot
parameters remain non-identifiable under permutation, zero-weight slots, and redistribution
of mass among atoms at the same location.

When $P\notin\mathcal A_{\leq K}$, the best projection need not be unique because
$\mathcal A_{\leq K}$ is nonconvex.  For example, under $D(x,y)=|x-y|$,
$P=(\delta_{-1}+\delta_1)/2$, and $K=1$, every $\delta_x$ with $x\in[-1,1]$ is optimal.
The result is pointwise in a fixed context; integrating over contexts yields the
corresponding conditional risk.  A shared neural model only searches its realizable
subfamily and nonconvex optimization need not reach the population optimum.  In this paper,
$D_z$ is a restricted Euclidean metric and, for $\beta>0$, $D_\beta$ is Euclidean distance
after a positive diagonal linear transformation, so both have strong negative type.  The unit-sphere latent
space is compact.  On the noncompact decoded Euclidean space, the identity and strict
identification remain valid for distributions with finite first $D_\beta$ moment, whereas
existence of an optimal atomic approximation additionally requires compactness or
coercivity.  The result would not support the same identification claim if these metrics
were replaced by squared Euclidean distance.

\section{Metrics}

\paragraph{Full-set scores.}
Trajectory Energy Score flattens each complete 60-step trajectory, divides it by
$\sqrt{60}$, and uses Euclidean distance; this is the root-mean-square displacement of the
whole block.  Endpoint Energy Score uses Euclidean endpoint distance.  All atom locations
and probabilities enter the observation and pairwise terms.  Router-top-1 selects
$\arg\max_k\pi_k(c)$ without access to the target; ties select the first stored index, so
the fixed-uniform arm deterministically reports head 0 as top-1.  Its ADE is the mean Euclidean
displacement over 60 future steps, FDE is terminal displacement, and MR is
$\mathbf 1\{\mathrm{FDE}>2\,\mathrm m\}$; ADE, FDE, and both decoded-space Energy Scores
are reported in meters.  Energy Score and expected ADE/FDE evaluate all atoms with their
probability mass, whereas router-top-1 evaluates one target-blind commitment.  A singleton
can have lower point error while giving worse distributional score and no alternative
support, so we report both views without treating either as a proxy for the other.

\paragraph{Router utilization.}
For scene $i$, $H(\pi_i)=-\sum_k\pi_{ik}\log\pi_{ik}$.  We compute
$\exp H(\pi_i)$ per scene and then average over validation scenes.  In contrast, an
``active head'' has at least $1\%$ of probability mass after aggregating router mass over
the complete validation set.

\paragraph{Maneuver events.}
An endpoint with displacement below 2\,m is \emph{stop}.  Otherwise its focal-frame angle is
left at $\geq30^\circ$, right at $\leq-30^\circ$, and straight between those thresholds.
Candidate probabilities are summed within the four bins.  Brier is the per-scene sum of four
squared probability errors.  ECE uses the maximum event probability and the correctness of
its event, aggregated over 10 equal-width confidence bins:
\[
\mathrm{ECE}=\sum_{b=1}^{10}\frac{n_b}{n}
\left|\operatorname{acc}(b)-\operatorname{conf}(b)\right|.
\]

\paragraph{Collision-adjusted latent support.}
The reported $N_{\mathrm{eff}}^{.10}$ is not a greedy clustering count.  For normalized
latents it is
\[
N_{\mathrm{eff}}^{.10}=
\left[\sum_{k,l}\pi_k\pi_l
\mathbf 1\{\lVert u_k-u_l\rVert_2\leq0.10\}\right]^{-1}.
\]
It equals one when all probability is carried by mutually duplicated atoms, discounts
near-duplicate pairs, and is invariant to atom ordering.  We compute it per validation scene
and then average over scenes.

\paragraph{Endpoint-deduplicated support.}
For the main official-validation matrix, we connect candidate endpoints whose Euclidean
distance is at most 1\,m, form the resulting connected components, sum router mass within
each component, and report the exponentiated entropy of the component masses.  This
quantity discounts endpoint-near-duplicate trajectories while preserving their total
probability mass.

\section{Per-Seed Results}

Tables~\ref{tab:joint-per-seed} and \ref{tab:anchor-per-seed} list every reported seed rather
than only aggregate means.  No seed or checkpoint was removed after inspecting these metrics.

\begin{table*}[t]
\centering
\scriptsize
\setlength{\tabcolsep}{3.2pt}
\begin{tabular}{@{}llrrrrrrr@{}}
\toprule
Arm & Seed & ES-traj. & ES-end. & Brier & ECE & traj. observation
& $\exp H(\pi)$ & active \\
\midrule
Point $K{=}1$ & 0 & 2.8387 & 5.5717 & .11942 & .05971 & 2.8387 & 1.000 & 1 \\
 & 1 & 2.8449 & 5.5842 & .12358 & .06179 & 2.8449 & 1.000 & 1 \\
 & 2 & 2.8493 & 5.5892 & .12078 & .06039 & 2.8493 & 1.000 & 1 \\
 & 3 & 2.8419 & 5.5742 & .11998 & .05999 & 2.8419 & 1.000 & 1 \\
 & 4 & 2.8381 & 5.5647 & .12102 & .06051 & 2.8381 & 1.000 & 1 \\
\midrule
Branch-JEPA / full-set, learned $\pi$ & 0 & 2.0755 & 4.0647 & .09224 & .01644 & 3.6217 & 5.779 & 6 \\
 & 1 & 2.0723 & 4.0479 & .09030 & .01309 & 3.6137 & 5.757 & 6 \\
 & 2 & 2.0760 & 4.0580 & .09198 & .01832 & 3.6246 & 5.813 & 6 \\
 & 3 & 2.0728 & 4.0575 & .09317 & .01774 & 3.6074 & 5.851 & 6 \\
 & 4 & 2.0864 & 4.0731 & .09200 & .01346 & 3.6251 & 5.786 & 6 \\
\midrule
Branch-JEPA / full-set, uniform $\pi$ & 0 & 2.1083 & 4.1182 & .09314 & .01485 & 3.6605 & 6.000 & 6 \\
 & 1 & 2.0996 & 4.1056 & .09248 & .01499 & 3.6270 & 6.000 & 6 \\
 & 2 & 2.1064 & 4.1214 & .09417 & .01387 & 3.6626 & 6.000 & 6 \\
 & 3 & 2.0969 & 4.1063 & .09410 & .01545 & 3.6428 & 6.000 & 6 \\
 & 4 & 2.1036 & 4.1143 & .09191 & .01472 & 3.6483 & 6.000 & 6 \\
\midrule
Branch-JEPA / soft-WTA & 0 & 2.1938 & 4.3041 & .09314 & .01272 & 3.9726 & 5.537 & 6 \\
 & 1 & 2.2083 & 4.3285 & .09319 & .01508 & 3.9953 & 5.509 & 6 \\
 & 2 & 2.2124 & 4.3429 & .09431 & .01499 & 3.9968 & 5.537 & 6 \\
 & 3 & 2.2110 & 4.3455 & .09386 & .01364 & 4.0077 & 5.526 & 6 \\
 & 4 & 2.1952 & 4.3058 & .09311 & .01220 & 3.9658 & 5.529 & 6 \\
\midrule
Branch-JEPA / partial Sinkhorn & 0 & 2.1910 & 4.3033 & .09338 & .01357 & 3.9612 & 5.505 & 6 \\
 & 1 & 2.2048 & 4.3233 & .09365 & .01336 & 3.9867 & 5.466 & 6 \\
 & 2 & 2.2236 & 4.3611 & .09458 & .01290 & 4.0178 & 5.474 & 6 \\
 & 3 & 2.2165 & 4.3501 & .09367 & .01245 & 4.0041 & 5.498 & 6 \\
 & 4 & 2.2000 & 4.3196 & .09410 & .01355 & 3.9753 & 5.503 & 6 \\
\midrule
Branch-JEPA / specialization & 0 & 2.2163 & 4.3430 & .09333 & .01242 & 4.0315 & 5.325 & 6 \\
 & 1 & 2.2140 & 4.3385 & .09293 & .01321 & 4.0340 & 5.288 & 6 \\
 & 2 & 2.2153 & 4.3413 & .09350 & .01169 & 4.0243 & 5.316 & 6 \\
 & 3 & 2.2355 & 4.3858 & .09445 & .01260 & 4.0699 & 5.312 & 6 \\
 & 4 & 2.2195 & 4.3521 & .09380 & .01248 & 4.0380 & 5.326 & 6 \\
\midrule
MDN / development-tuned & 0 & 2.3438 & 4.6042 & .09972 & .02545 & 3.9742 & 3.783 & 6 \\
 & 1 & 2.5272 & 4.9753 & .10683 & .03201 & 4.2579 & 3.810 & 6 \\
 & 2 & 2.3157 & 4.5469 & .09883 & .02036 & 4.0460 & 4.064 & 6 \\
 & 3 & 2.3475 & 4.6108 & .10024 & .02454 & 3.9967 & 3.841 & 6 \\
 & 4 & 2.3796 & 4.6775 & .10086 & .02517 & 4.0275 & 3.775 & 6 \\
\bottomrule
\end{tabular}
\vspace{2mm}

\textbf{Matched-$K$ aggregate expected and top-1 point metrics}\\[1mm]
\setlength{\tabcolsep}{4.6pt}
\begin{tabular}{@{}lrrrrr@{}}
\toprule
Arm & expected ADE & expected FDE & top-1 ADE & top-1 FDE & top-1 MR\\
\midrule
Branch-JEPA / full-set
& $2.7941{\pm}.0059$ & $7.1904{\pm}.0170$
& $2.4694{\pm}.0888$ & $6.2939{\pm}.2270$ & $.7236{\pm}.0166$\\
Branch-JEPA / full-set, uniform
& $2.8214{\pm}.0108$ & $7.2293{\pm}.0410$
& $2.7383{\pm}.4421$ & $7.0138{\pm}1.2861$ & $.7795{\pm}.0608$\\
Branch-JEPA / soft-WTA
& $3.0677{\pm}.0139$ & $7.9442{\pm}.0347$
& $2.4858{\pm}.0223$ & $6.3828{\pm}.0564$ & $.7014{\pm}.0018$\\
Branch-JEPA / partial Sinkhorn
& $3.0695{\pm}.0192$ & $7.9448{\pm}.0363$
& $2.5143{\pm}.0117$ & $6.4523{\pm}.0322$ & $.7056{\pm}.0018$\\
Branch-JEPA / specialization
& $3.1103{\pm}.0133$ & $8.0348{\pm}.0360$
& $2.4932{\pm}.0125$ & $6.3876{\pm}.0350$ & $.7018{\pm}.0016$\\
MDN / development-tuned
& $3.1253{\pm}.0868$ & $8.0942{\pm}.2273$
& $2.5609{\pm}.0860$ & $6.5957{\pm}.2245$ & $.7179{\pm}.0084$\\
\bottomrule
\end{tabular}
\vspace{1.2mm}

\textbf{Paired full-set minus specialization statistics}\\[.5mm]
\setlength{\tabcolsep}{6.0pt}
\begin{tabular}{@{}lrrc@{}}
\toprule
Metric & mean difference & paired-$t$ 95\% interval & common direction \\
\midrule
Trajectory ES & $-.1435$ & $[-.1574,-.1295]$ & 5/5 seeds \\
Trajectory observation & $-.4211$ & $[-.4512,-.3909]$ & 5/5 seeds \\
Expected ADE (m) & $-.3161$ & $[-.3392,-.2930]$ & 5/5 seeds \\
\bottomrule
\end{tabular}
\caption{Every locked seed on the 24,988-scene official AV2 validation split, plus the
parameter-matched $K{=}1$ structural control, followed by aggregate expected and
router-top-1 point metrics for the six $K{=}6$ rows.  Distances are in meters.  The
trajectory observation term is ES-traj plus one half of probability-weighted pairwise
trajectory diversity.  Every configuration and checkpoint was selected on the disjoint
train-derived development split before this one-shot evaluation; no official-split
retuning, seed removal, or checkpoint replacement was performed.}
\label{tab:joint-per-seed}
\end{table*}

Table~\ref{tab:joint-per-seed} includes the three prespecified paired
full-set-minus-specialization comparisons; each has the same direction in all five
seeds.  Router-top-1 ADE also has the best \mbox{matched-$K$} mean for full-set
training; complete paired intervals for every retained metric and comparator
accompany the released aggregate table.

\begin{table*}[t]
\centering
\scriptsize
\setlength{\tabcolsep}{3.8pt}
\begin{tabular}{@{}llrrrrr@{}}
\toprule
Branch location & Seed & ES-traj. $\downarrow$ & expected ADE $\downarrow$
& Brier $\downarrow$ & $\exp H(\pi)$ $\uparrow$ & top-1 ADE $\downarrow$\\
\midrule
Latent (Branch-JEPA) & 0 & 2.07554 & 2.79683 & .092245 & 5.7790 & 2.44859\\
 & 1 & 2.07235 & 2.79085 & .090297 & 5.7571 & 2.36108\\
 & 2 & 2.07605 & 2.79954 & .091981 & 5.8128 & 2.60835\\
 & 3 & 2.07281 & 2.78544 & .093169 & 5.8512 & 2.46677\\
 & 4 & 2.08639 & 2.79807 & .092000 & 5.7864 & 2.46198\\
\midrule
Output only (one latent) & 0 & 2.11748 & 2.83616 & .095338 & 5.1814 & 2.46605\\
 & 1 & 2.11664 & 2.83760 & .096117 & 5.2917 & 2.44981\\
 & 2 & 2.10778 & 2.83739 & .095643 & 5.2547 & 2.43477\\
 & 3 & 2.11105 & 2.82435 & .095644 & 5.2312 & 2.41962\\
 & 4 & 2.11807 & 2.81976 & .094942 & 5.3120 & 2.46466\\
\midrule
Latent mean$\pm$SD & -- & $2.07663{\pm}.00569$ & $2.79414{\pm}.00588$
& $.091938{\pm}.001038$ & $5.7973{\pm}.0361$ & $2.4694{\pm}.0888$\\
Output mean$\pm$SD & -- & $2.11421{\pm}.00455$ & $2.83105{\pm}.00839$
& $.095537{\pm}.000434$ & $5.2542{\pm}.0514$ & $2.4470{\pm}.0199$\\
\bottomrule
\end{tabular}
\caption{\textbf{Latent branching versus output-only branching on official validation.}
Both emit six weighted trajectories, use exactly
$3{,}608{,}958$ downstream trainable parameters, the same five seeds and 30-epoch
budget, and target-blind inference.  Checkpoints were selected on the train-derived
development split and evaluated once without official-split tuning.  Branch-JEPA improves
ES, expected ADE, Brier, and mass-effective support in all five paired seeds.  The top-1
ADE difference is unresolved; this control isolates finite-support representation rather
than a universal point-error gain.}
\label{tab:latent-vs-output-official}
\end{table*}

For latent minus output-only branching, paired mean differences are
$-.03758$ trajectory ES (paired-$t$ 95\% interval
$[-.04475,-.03041]$), $-.03691$\,m expected ADE
($[-.04834,-.02547]$), $-.003598$ Brier
($[-.005228,-.001969]$), and $+.5431$ effective modes
($[+.4556,+.6306]$).  All four have the same direction in all five seeds.
The top-1 ADE difference is $+.0224$\,m with interval
$[-.0987,+.1435]$ and is not used as evidence of an advantage.

\begin{figure*}[!t]
\centering
\includegraphics[width=.98\textwidth]{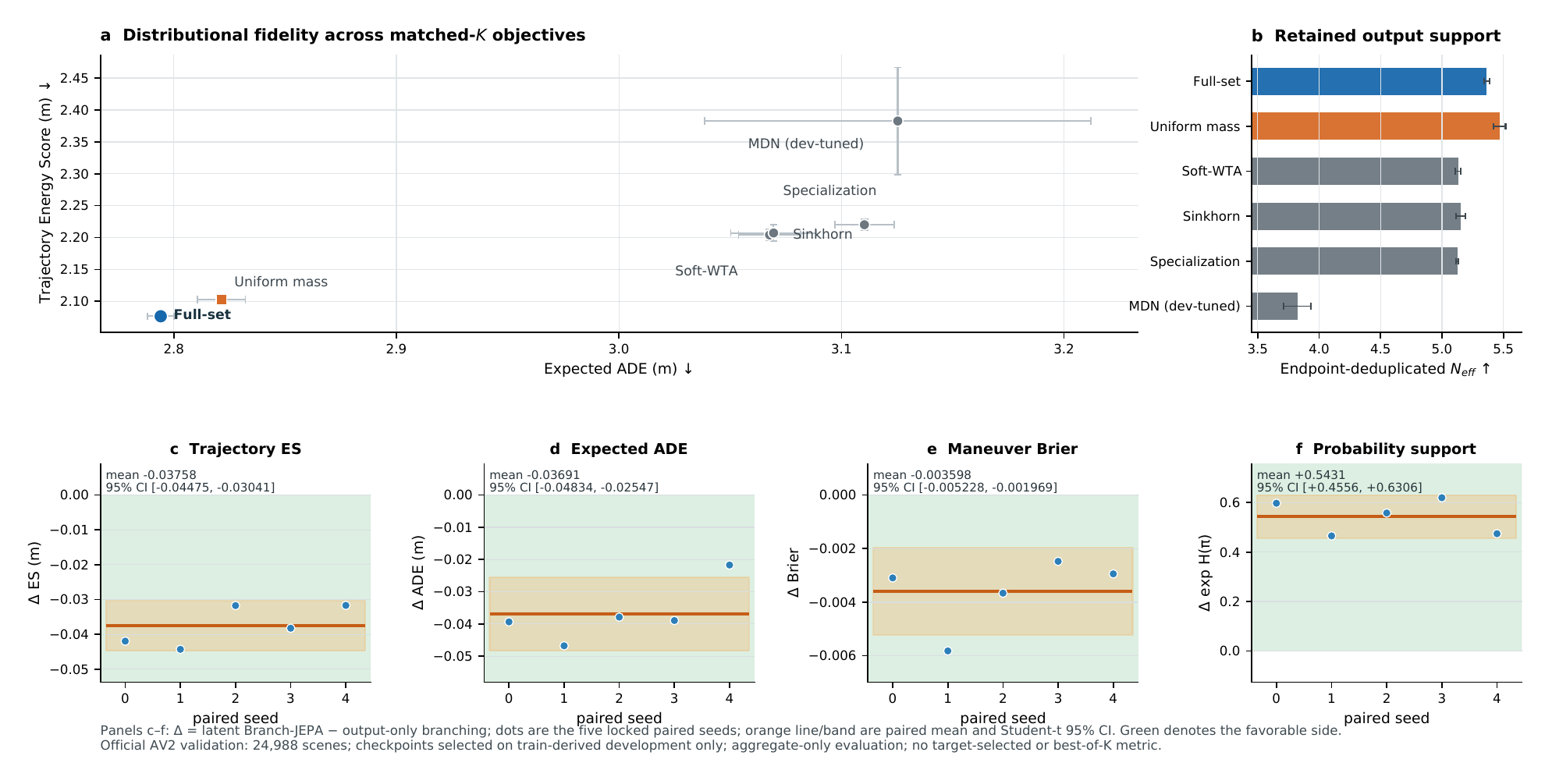}
\caption{\textbf{Official-validation distribution and architecture summary.}
Panels (a)--(b) compare all six locked, matched-$K{=}6$ objectives on the
24,988-scene official AV2 validation split.  Lower and farther left is better in
(a); (b) reports endpoint-deduplicated effective support rather than nominal head
count.  Panels (c)--(f) show every paired seed difference for latent
Branch-JEPA minus the parameter-exact output-only control; dots are the five locked
seeds and orange bands are paired-$t$ 95\% intervals.  The two support panels use
their explicitly labeled registered summaries: endpoint-deduplicated support for
the objective matrix and probability support $\exp H(\pi)$ for the architecture
isolation pair.  Checkpoints were selected only on the disjoint train-derived
development split.  No panel uses a target-selected candidate or best-of-$K$
metric.}
\label{fig:official-validation-summary-supp}
\end{figure*}

\begin{table*}[t]
\centering
\scriptsize
\setlength{\tabcolsep}{4.2pt}
\begin{tabular}{@{}lcccclrrrr@{}}
\toprule
control & $\lambda_z$ & $\lambda_y$ & $\lambda_{\rm rec}$ & target signal
& role & ES-traj. $\downarrow$ & expected ADE $\downarrow$
& Brier $\downarrow$ & $\exp H(\pi)$ $\uparrow$\\
\midrule
Decoded set only & 0 & 1 & 0 & no & target-path ablation
& $2.05772{\pm}.00819$ & $2.79041{\pm}.01284$
& $.09465{\pm}.00163$ & $5.593{\pm}.042$\\
Latent + decoded & 1 & 1 & 0 & yes & reconstruction ablation
& $\mathbf{2.04925{\pm}.01067}$ & $2.76141{\pm}.01288$
& $.09354{\pm}.00080$ & $5.789{\pm}.017$\\
Full objective & 1 & 1 & 1 & yes & paper setting
& $2.05603{\pm}.00329$ & $2.76438{\pm}.01326$
& $.09328{\pm}.00141$ & $\mathbf{5.806{\pm}.010}$\\
\midrule
$\lambda_z=.5$ & .5 & 1 & 1 & yes & latent-weight sweep
& $2.05930{\pm}.00707$ & $2.78107{\pm}.01974$
& $.09431{\pm}.00104$ & $5.799{\pm}.038$\\
$\lambda_z=2$ & 2 & 1 & 1 & yes & latent-weight sweep
& $2.05395{\pm}.01221$ & $\mathbf{2.76396{\pm}.02129}$
& $\mathbf{.09269{\pm}.00139}$ & $5.771{\pm}.035$\\
$\lambda_y=.5$ & 1 & .5 & 1 & yes & decoded-weight sweep
& $2.05765{\pm}.01104$ & $2.76998{\pm}.01496$
& $.09335{\pm}.00071$ & $5.760{\pm}.055$\\
$\lambda_y=2$ & 1 & 2 & 1 & yes & decoded-weight sweep
& $2.05459{\pm}.00530$ & $2.77643{\pm}.01912$
& $.09403{\pm}.00132$ & $5.766{\pm}.018$\\
\bottomrule
\end{tabular}
\caption{\textbf{Five-seed loss and weight ablation on the train-derived development
split.}  All rows keep the $K{=}6$ latent-branch architecture, initialization protocol,
and 30-epoch budget.  The decoded-only row removes all target-path gradients, although
the inert EMA tower remains instantiated.  Relative to decoded-only, the full objective
improves expected ADE and effective support in every paired seed; the corresponding
full-minus-control 95\% intervals are $[-.0476,-.0044]$ and $[.155,.271]$.
Trajectory-ES and Brier intervals overlap zero.  The weight sweep shows that the result is
not sensitive to a narrow $\lambda_z/\lambda_y$ choice.}
\label{tab:loss-ablation}
\end{table*}

\begin{table*}[t]
\centering
\footnotesize
\setlength{\tabcolsep}{4.2pt}
\begin{tabular}{@{}llrrrrrrr@{}}
\toprule
Arm & Seed & latent ES & traj. ES (m) & end. ES (m) & Brier & ECE & top-1 FDE (m)
& $N_{\mathrm{eff}}^{.10}$ \\
\midrule
Point $K{=}1$ & 0 & .21030 & 2.8843 & 5.6710 & .12142 & .06071 & 5.6710 & 1.000 \\
Point $K{=}1$ & 1 & .20228 & 2.8769 & 5.6540 & .12230 & .06115 & 5.6540 & 1.000 \\
Point $K{=}1$ & 2 & .21152 & 2.9147 & 5.7258 & .12142 & .06071 & 5.7258 & 1.000 \\
\midrule
Branch-JEPA / full-set $K{=}6$ & 0 & .15431 & 2.1286 & 4.1778 & .09387 & .01557 & 6.3528 & 4.241 \\
Branch-JEPA / full-set $K{=}6$ & 1 & .14838 & 2.1171 & 4.1459 & .09074 & .01232 & 6.2698 & 4.220 \\
Branch-JEPA / full-set $K{=}6$ & 2 & .15469 & 2.1454 & 4.2112 & .09329 & .01429 & 6.4211 & 4.319 \\
\bottomrule
\end{tabular}
\caption{Per-seed frozen-anchor Stage-B results.  Target encoder and decoder drift are
exactly zero in every run.  The $K{=}1$ rows have lower top-1 FDE in every seed, while
the Branch-JEPA rows improve full-distribution scores and retain collision-adjusted
support above four; these metrics answer different prediction questions.}
\label{tab:anchor-per-seed}
\end{table*}

\section{Frozen-Anchor Gates}

Stage A trains a future-view encoder and normalized-latent decoder with
\begin{equation}
\begin{aligned}
\mathcal L_A&=\operatorname{SmoothL1}(d(\bar z),Y^+)
+\mathcal L_{\mathrm{var}}+.25\,\mathcal L_{\mathrm{rank}},\\
\mathcal L_{\mathrm{var}}&=\frac{1}{d_z}\sum_j
\left[.02-\sqrt{\operatorname{Var}_{\mathcal B}[\bar z_j]+10^{-4}}\right]_+,\\
\mathcal L_{\mathrm{rank}}&=\frac{[8-r_{\mathrm{PR}}]_+}{8},\qquad
r_{\mathrm{PR}}=\frac{\operatorname{tr}(C)^2}{\lVert C\rVert_F^2}.
\end{aligned}
\label{eq:anchor-objective}
\end{equation}
where $\bar z$ is the normalized target latent and $C$ its minibatch covariance.  Latent
normalization and covariance penalties are computed in FP32.  The rank-floor regularizer
therefore uses the minibatch participation ratio $r_{\mathrm{PR}}$.  The checkpoint gate
instead uses the full-validation covariance eigenvalues $\lambda_j$ and entropy effective
rank
\[
q_j=\frac{\lambda_j}{\sum_\ell\lambda_\ell},\qquad
r_{\mathrm{ent}}=\exp\!\left(-\sum_j q_j\log q_j\right).
\]
These are distinct summaries; $r_{\mathrm{PR}}$ is the exponentiated order-2 R\'enyi entropy
and is a conservative lower bound on $r_{\mathrm{ent}}$ for a common covariance spectrum.
A checkpoint is eligible only if
decoder ADE/FDE are at most $2/4$\,m, mean normalized dimension standard deviation is at
least $.005$, the fraction below $10^{-3}$ is at most $.90$, and entropy effective rank is
at least 8.  Among eligible epochs, minimum validation decoder ADE selects the checkpoint;
the thresholds were fixed before the formal three-seed run.

\begin{table*}[t]
\centering
\small
\setlength{\tabcolsep}{3.5pt}
\begin{tabular}{@{}lrrrrrl@{}}
\toprule
Seed & ADE & FDE & mean std & low-std frac. & eff. rank & SHA-256 prefix \\
\midrule
0 & .1827 & .3216 & .02187 & 0 & 12.911 & \texttt{7721aad75852} \\
1 & .1781 & .2918 & .02108 & 0 & 12.860 & \texttt{8585e6e7b7cf} \\
2 & .1841 & .3192 & .02184 & 0 & 13.196 & \texttt{a797ffe511bb} \\
\bottomrule
\end{tabular}
\caption{Accepted Stage-A anchors.  All selected checkpoints occur at epoch 30.  Passing
these numerical gates rules out the measured collapse modes but does not prove that the
representation preserves every semantic distinction.}
\label{tab:anchor-gates}
\end{table*}

\section{Official Raw-Map Audits}

The locked primary evaluator also runs a prespecified official-polygon summary on the
official validation split.  Across the five Branch-JEPA seeds, probability-weighted
candidate-point containment is $99.319{\pm}.041\%$, the probability mass of trajectories
with at least 95\% contained center points is $97.157{\pm}.181\%$, and the
mass-weighted feasible effective support is $5.399{\pm}.025$.  These are
mean${\pm}$sample SD across seed-level aggregates and establish center-point containment,
not lane connectivity or dynamic safety.

Separately, for the frozen-anchor attribution audit, we export all six trajectories and
router probabilities from every checkpoint,
restore them to world coordinates, and evaluate the 22,979 validation scenes whose focal
actor is VEHICLE or BUS.  The realized future is not loaded.  Predictions are neither
projected nor map-matched.  Geometry is the union of official AV2 drivable-area polygons
with 0.1\,m boundary tolerance; $22{,}977/22{,}979$ observed focal origins fall inside this
union, providing a coordinate-transform sanity check.

A candidate is feasible when at least 95\% of its 60 center points are inside.  Feasible
candidates are connected into duplicate components when their trajectory RMS distance is at
most 0.5\,m.  Component probabilities are summed.  Unique support counts components with
probability at least $.01$.  Feasible entropy support is the exponentiated entropy of
feasible component mass after renormalization; it is set to zero for a scene with no feasible
candidate, and the reported value averages over all scenes.  This zero convention explains
why the point-model mean can be below one.  The mass-weighted version multiplies this count
by total feasible probability mass.  The diversity quantities below also retain the
original router mass.  For feasibility indicators $f_k$, trajectory RMS distance
$D^{\mathrm{traj}}_{kl}$, and endpoint distance $D^{\mathrm{end}}_{kl}$, we report
\[
\begin{aligned}
\mathrm{Div}_{\mathrm{traj}}&=\sum_{k,l}\pi_k\pi_l f_kf_l
D^{\mathrm{traj}}_{kl},\\
\mathrm{Div}_{\mathrm{end}}&=\sum_{k,l}\pi_k\pi_l f_kf_l
D^{\mathrm{end}}_{kl}.
\end{aligned}
\]
Thus they decrease when probability leaves the feasible set; they are not conditionally
renormalized diversity scores.

\begin{table*}[t]
\centering
\footnotesize
\setlength{\tabcolsep}{4.2pt}
\begin{tabular}{@{}lccc@{}}
\toprule
Metric & Standard JEPA $K{=}1$ & Branch-JEPA / full-set $K{=}6$ & Branch$-$Standard [95\% CI] \\
\midrule
Drivable point mass (\%) & $99.577{\pm}.027$ & $99.220{\pm}.043$
& $-.358$ [$-.386,-.329$] \\
Drivable trajectory mass (\%) & $98.165{\pm}.064$ & $96.775{\pm}.153$
& $-1.390$ [$-1.493,-1.286$] \\
Feasible unique support & $.982{\pm}.001$ & $5.591{\pm}.018$
& $+4.609$ [$4.600,4.617$] \\
Feasible entropy support (empty${=}0$) & $.982{\pm}.001$ & $5.205{\pm}.023$
& $+4.223$ [$4.212,4.233$] \\
Mass-weighted feasible effective support & $.982{\pm}.001$ & $5.093{\pm}.027$
& $+4.111$ [$4.098,4.123$] \\
Mass-weighted feasible trajectory diversity (m) & 0 & $3.149{\pm}.030$
& $+3.149$ [$3.133,3.164$] \\
Mass-weighted feasible endpoint diversity (m) & 0 & $6.426{\pm}.065$
& $+6.426$ [$6.393,6.458$] \\
\bottomrule
\end{tabular}
\caption{Official-map audit (mean and population SD across three seed means).
Intervals are paired-scene-bootstrap intervals conditional on those fitted models.}
\label{tab:raw-map-full}
\end{table*}

\begin{figure*}[!t]
\centering
\includegraphics[width=.98\textwidth]{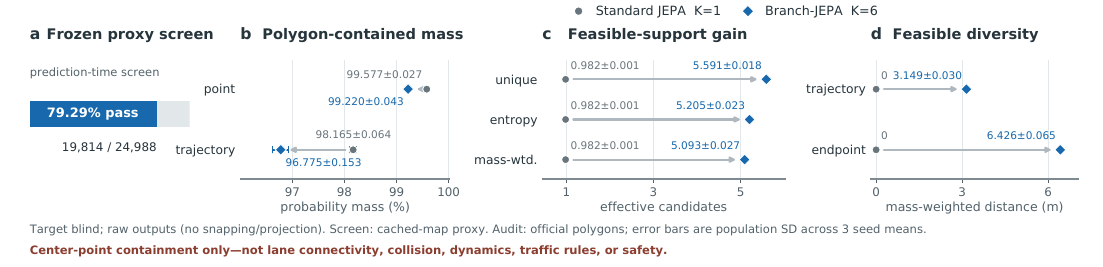}
\caption{\textbf{Target-blind AV2 proxy and official raw-map audit.}
(a) Frozen cached-map screen; (b) official-polygon point and trajectory mass;
(c) feasible effective-support measures; and (d) feasible trajectory/endpoint
diversity.  Arrows run from Standard JEPA ($K{=}1$) to Branch-JEPA ($K{=}6$),
and labels are mean${\pm}$population SD across three seed means.}
\label{fig:av2-map-audit-summary-supp}
\end{figure*}

The audit establishes substantial polygon-contained support, not non-inferior compliance:
full-set Branch-JEPA's drivable-trajectory mass is 1.39 percentage points lower.  Center-point
containment also does not test directed lane connectivity, actor footprint, collision,
dynamics, traffic rules, or closed-loop safety.

\begin{figure*}[!t]
\centering
\includegraphics[width=.82\textwidth]{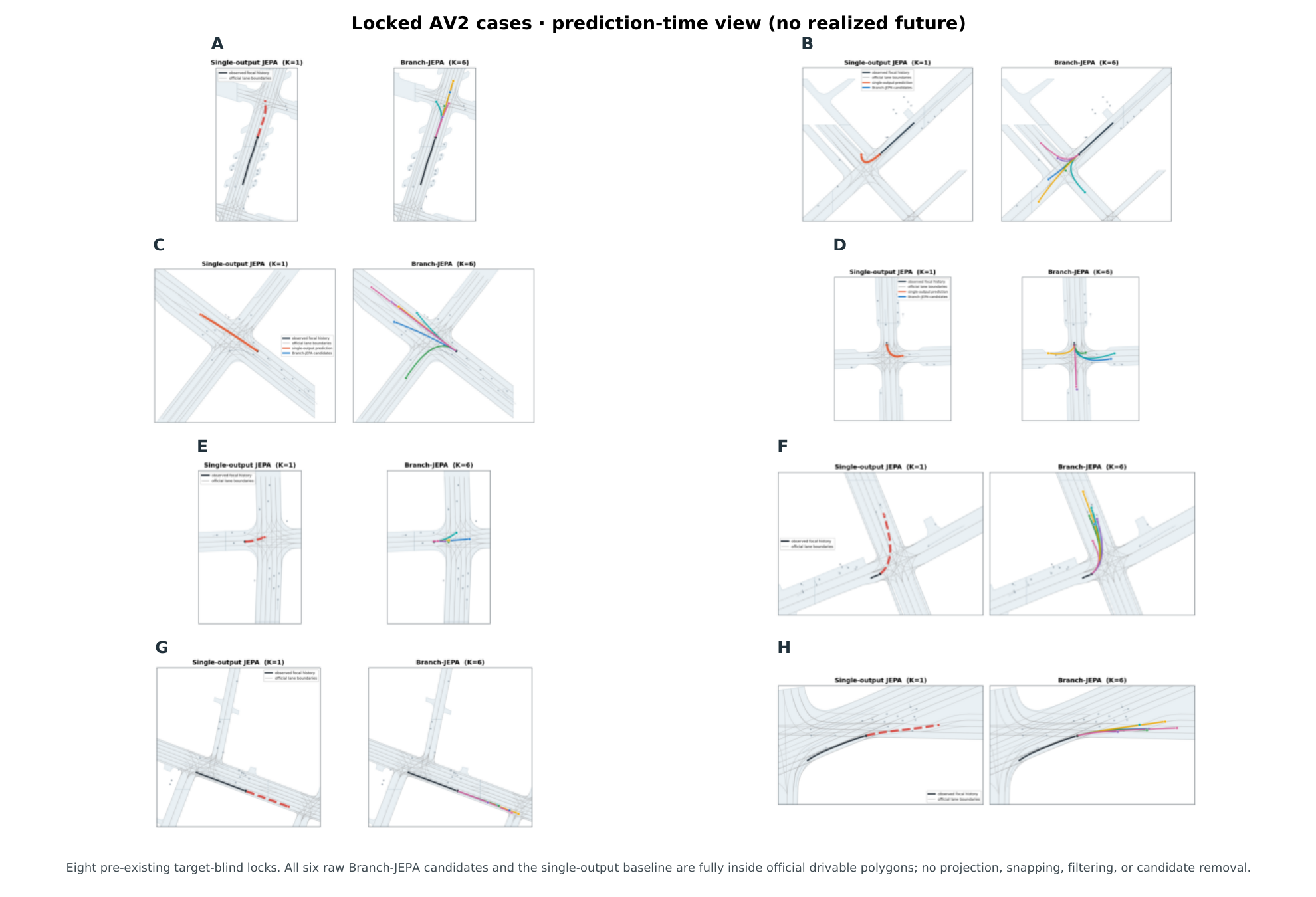}\\[-1mm]
\includegraphics[width=.82\textwidth]{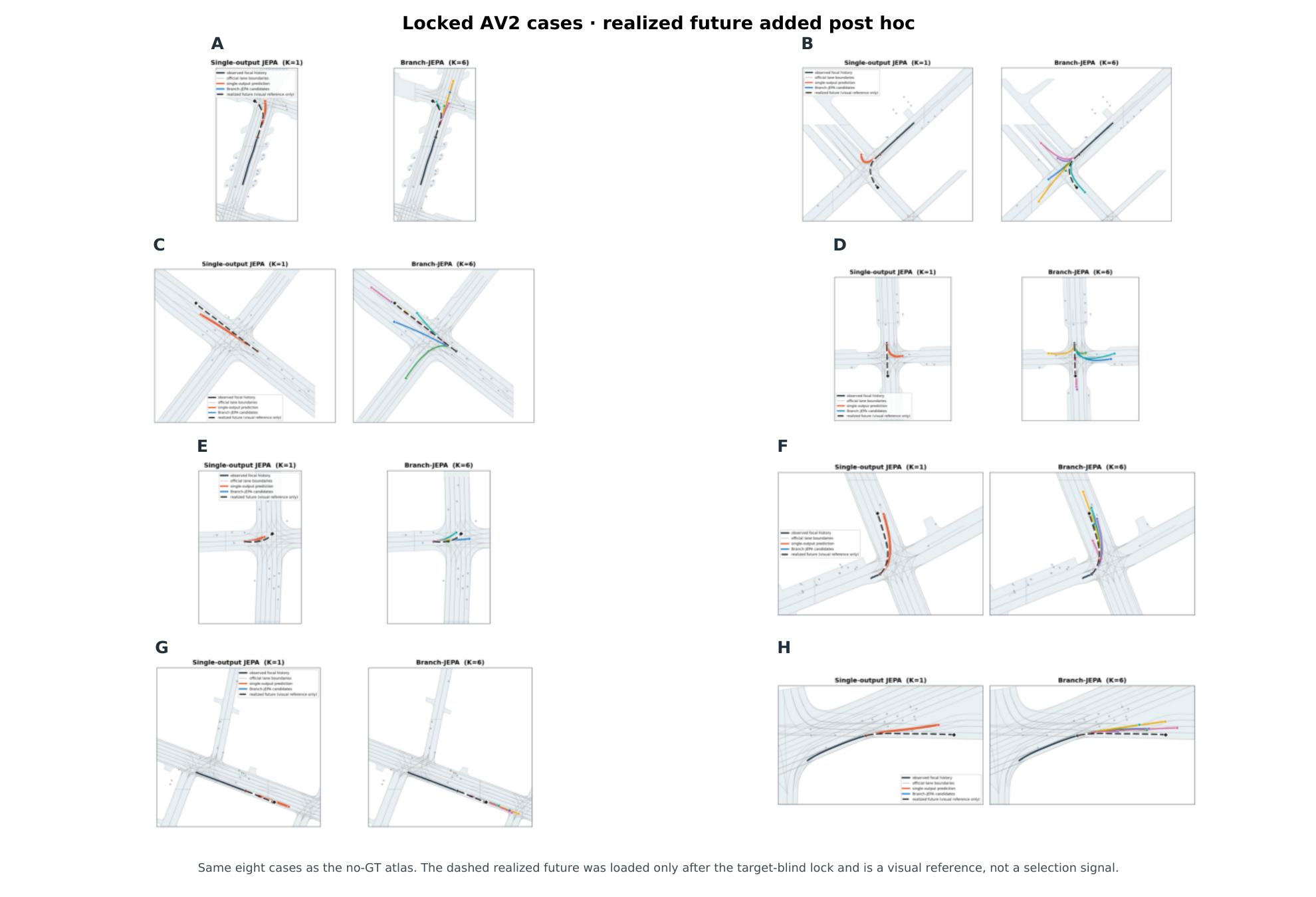}
\caption{\textbf{Paired locked AV2 atlas.}
Top: the prediction-time view of all eight pre-existing target-blind locks, containing
the raw single path and all six raw Branch-JEPA paths.  Bottom: the identical cases,
maps, and predictions with only the dashed realized future added after locking.
Every displayed trajectory has 100\% center-point containment in official drivable
polygons; the bottom panel is a post-hoc diagnostic, not a selection signal.}
\label{fig:av2-atlas-no-gt-supp}
\label{fig:av2-atlas-posthoc-supp}
\end{figure*}

\section{Target-Blind Qualitative Protocol}

Figure~2 in the main paper is a mechanism illustration, not a quantitative validity test.
Its frozen prediction-time screen accesses observed focal and neighbor histories, the 48
cached lane/crosswalk polylines, official static-map polygons, \mbox{scenario identifiers}, and
model outputs only.  It never loads the realized future, a future-selected candidate
index, or an error to the target.  Predictions are rendered unchanged: there is no map
snapping, projection, candidate deletion, or probability renormalization.

For the parent gallery, a candidate point is within the cached-map proxy when it is at most
3\,m from one of the cached polylines.  A context passes when its probability-weighted
outside-point fraction is at most $.10$ and at least $.80$ router mass is carried by
candidates satisfying that threshold.  This frozen target-blind screen passes
$19{,}814/24{,}988=79.29\%$ of validation contexts.  The six displayed contexts satisfy
the stricter conditions that every candidate has zero points beyond the 3\,m proxy and
that every raw candidate point is inside an official drivable polygon.  Their identities,
raw-map hashes, per-candidate containment values, and source-image hashes are fixed in the
archived figure manifest before the realized future is added for reference.  This
case-level check and the population-level official-polygon audit in
Table~\ref{tab:raw-map-full} establish center-point containment only; neither guarantees
lane connectivity nor that the displayed alternatives cover the realized future.

\subsection{Extended AV2 prediction atlas}

Figure~\ref{fig:av2-atlas-no-gt-supp} shows the same eight pre-existing target-blind
locks twice in the same order.  The top panel is exactly the prediction-time view:
observed history, official map geometry, the single-output prediction, and all six
raw Branch-JEPA candidates.  The bottom panel changes only one layer by adding
the realized future after the lock.  Thus a reader can first inspect the model's
enumerated alternatives without outcome information and then use the lower panel
as a post-hoc diagnostic.  Every displayed raw trajectory has 100\% center-point
containment in the official drivable polygons; none was snapped, projected,
filtered, removed, or selected by target error.

\paragraph{Atlas audit key.}
The upper panel contains prediction-time information only; the lower panel adds the
future after locking.  Both retain the unchanged single path and all six Branch-JEPA
paths for every case.

\section{Cross-Domain Interface Diagnostics}

The following experiments evaluate Branch-JEPA with specialization training.  They test
whether the shared multi-branch architecture exposes context-dependent successor support
across observation regimes; the AV2 experiments separately compare specialization with
full-set training.

\begin{table*}[!t]
\centering
\footnotesize
\textbf{(a) planAll: independent graph-fit campaign (\%)}\\[2pt]
\setlength{\tabcolsep}{5.0pt}
\begin{tabular}{@{}lccc@{}}
\toprule
predictor & medium & teleport & large\\
\midrule
Standard JEPA & $8.4{\pm}3.0$ & $3.7{\pm}1.0$ & $1.6{\pm}0.7$\\
fused MoE control & $7.5{\pm}1.7$ & $3.6{\pm}0.9$ & $3.1{\pm}1.2$\\
variational control & $8.5{\pm}1.6$ & $3.5{\pm}1.6$ & $3.3{\pm}0.4$\\
MDN & $56.6{\pm}22.6$ & $51.2{\pm}13.4$ & $\mathbf{71.7{\pm}13.7}$\\
\textbf{Branch-JEPA}
& $\mathbf{85.1{\pm}8.7}$ & $\mathbf{74.8{\pm}9.6}$ & $38.6{\pm}50.0$\\
codebook control$^\dagger$ & $26.7{\pm}2.7$ & $16.7{\pm}1.9$ & $14.9{\pm}4.0$\\
\bottomrule
\end{tabular}
\vspace{5pt}

\scriptsize
\textbf{(b) \texttt{valid2}: verified route and macro transition precision (\%)}\\[2pt]
\setlength{\tabcolsep}{3.1pt}
\begin{tabular}{@{}lcccccc@{}}
\toprule
& \multicolumn{3}{c}{verified route $\uparrow$}
& \multicolumn{3}{c}{macro transition precision $\uparrow$}\\
\cmidrule(lr){2-4}\cmidrule(lr){5-7}
predictor & medium & teleport & large & medium & teleport & large\\
\midrule
Standard JEPA
& $8.3{\pm}2.9$ & $3.5{\pm}0.9$ & $1.6{\pm}0.7$
& $90.0{\pm}5.8$ & $81.8{\pm}3.6$ & $62.2{\pm}7.8$\\
MDN
& $13.4{\pm}3.7$ & $3.9{\pm}1.2$ & $5.7{\pm}3.0$
& $21.1{\pm}3.3$ & $18.4{\pm}5.9$ & $14.1{\pm}1.5$\\
\textbf{Branch-JEPA}
& $\mathbf{24.7{\pm}8.7}$ & $\mathbf{19.2{\pm}10.4}$ & $\mathbf{13.9{\pm}16.7}$
& $49.2{\pm}5.5$ & $54.1{\pm}22.6$ & $56.1{\pm}7.4$\\
\bottomrule
\end{tabular}
\vspace{6pt}

\scriptsize
\textbf{(c) Separate higher-dimensional raw-support campaigns (\%)}\\[2pt]
\setlength{\tabcolsep}{3.0pt}
\begin{tabular}{@{}lrrrr@{}}
\toprule
observation / metric $\uparrow$ & Standard & fused & MDN
& Branch-JEPA\\
\midrule
29-D AntMaze / raw COV-R
& $29.4{\pm}1.0$ & $24.2{\pm}0.4$ & $38.6{\pm}1.0$
& $\mathbf{57.6{\pm}2.3}$\\
RGB AntMaze / raw COV-R
& $24.3{\pm}1.2$ & $27.0{\pm}0.7$ & $26.7{\pm}2.7$
& $\mathbf{61.3{\pm}12.1}$\\
\bottomrule
\end{tabular}
\vspace{6pt}

\includegraphics[width=.96\textwidth]{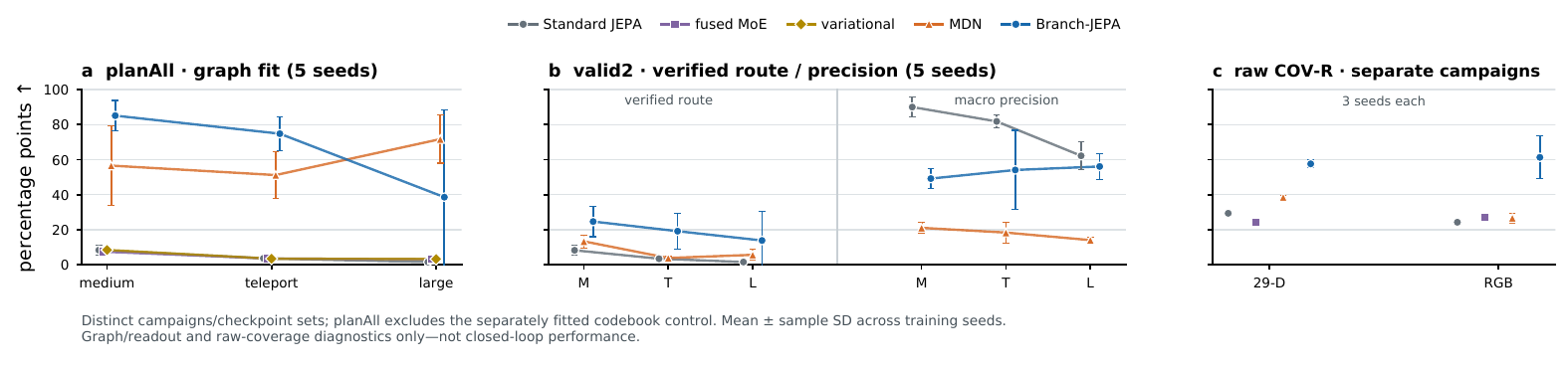}\\[-1mm]
\textbf{\scriptsize(d) Campaign-separated visual summary}
\caption{\textbf{Campaign-separated OGBench diagnostics.}
Panels (a)--(c) report the independent graph-fit, \texttt{valid2}, and
higher-dimensional campaigns; (d) visualizes the same campaign-separated aggregates.
The $^\dagger$codebook is separately fitted, and all values are mean${\pm}$sample SD
across training seeds in percentage points.  Campaigns do not share checkpoints.}
\label{tab:interface-benchmarks-supp}
\end{table*}

\paragraph{OGBench information and readout protocols.}
Table~\ref{tab:interface-benchmarks-supp} combines distinct, artifact-backed campaigns and is not a set of measurements
of one common checkpoint set.  The planAll campaign fits action-marginalized point-maze
predictors to current coordinates; its learned arms and separately fitted codebook
control appear in panel (a).  A second five-seed retraining campaign
(\texttt{valid2}) contains only Standard JEPA, MDN, and Branch-JEPA and supplies both
verified-route existence and transition precision in panel (b).  Both graph campaigns
use $K{=}8$, 30k updates, 512-dimensional latents, coordinate features, and the same
edge-decoding rule, but their trained checkpoints differ.

For these graph diagnostics, $20\%$ of unique directed edges are withheld from fitting
subject to retaining source-cell training support.  A candidate creates an edge when its
cosine similarity to a cell latent exceeds $.5$; all heads are retained.  planAll is the
success rate on up to 300 sampled, truly reachable start--goal queries under
predicted-graph BFS, not exhaustive all-pairs evaluation.  Verified-route existence uses
a separate query set, retains only predicted edges that are real transitions in the full
environment graph, and asks whether a complete path remains.  Transition precision is the
macro-average of per-source fractions of proposed edges that are real.

The higher-dimensional campaigns in panel (c) use still other checkpoints and report
raw candidate coverage only.  They are not interchangeable with the graph diagnostics,
and none of these quantities is closed-loop task success.

\begin{figure*}[!t]
\centering
\includegraphics[width=0.82\textwidth]{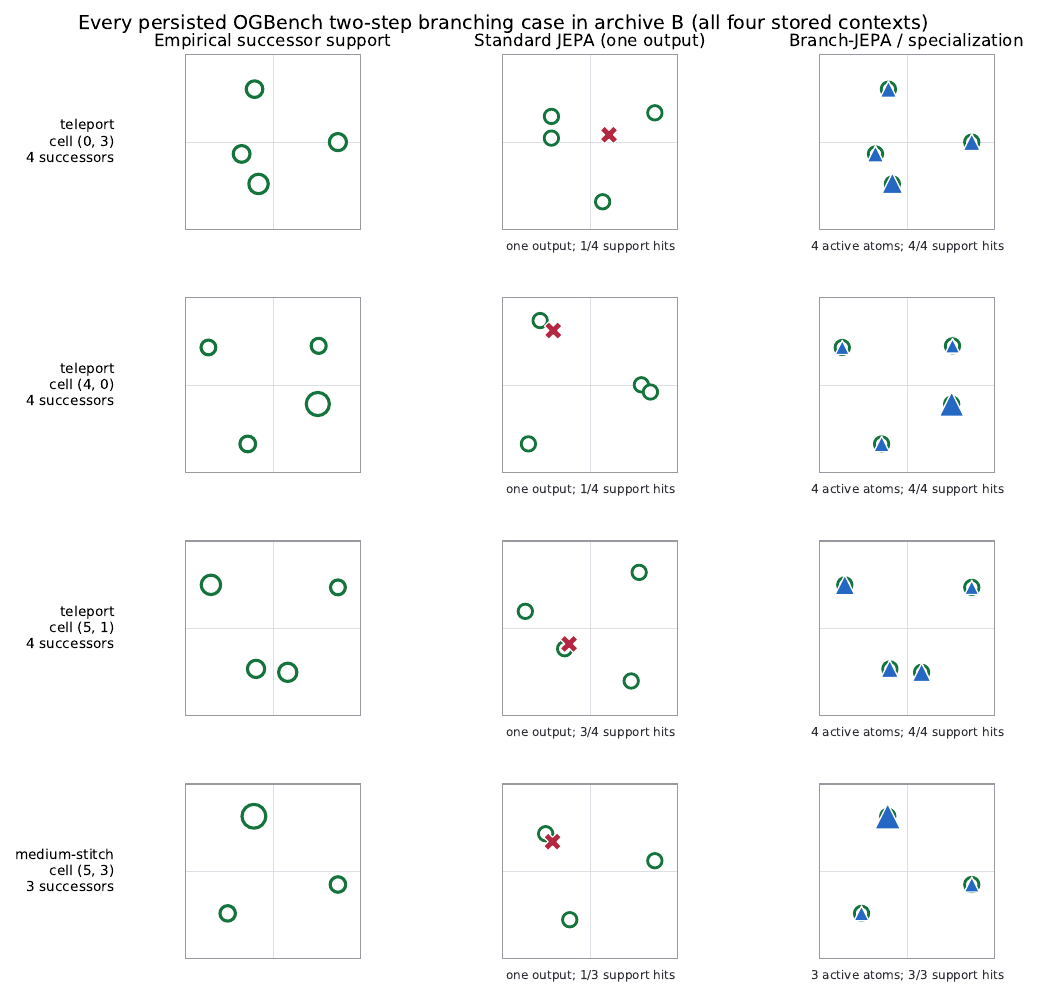}\\[-1mm]
\textbf{\small(a) Every context in two-step archive B}\\[1mm]
\begin{minipage}[t]{0.66\textwidth}
\centering
\includegraphics[width=.92\linewidth]{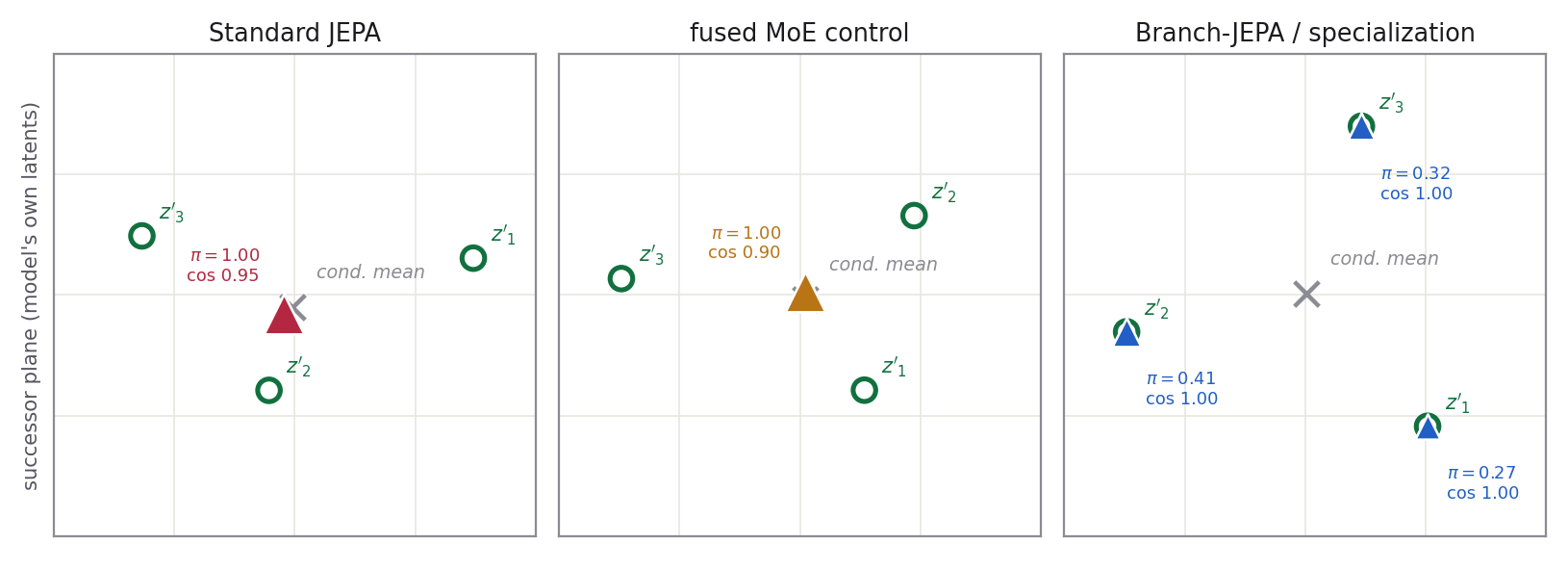}\\[-1mm]
\textbf{\small(b) Trained successor geometry}
\end{minipage}\hfill
\begin{minipage}[t]{0.33\textwidth}
\centering
\includegraphics[width=.92\linewidth]{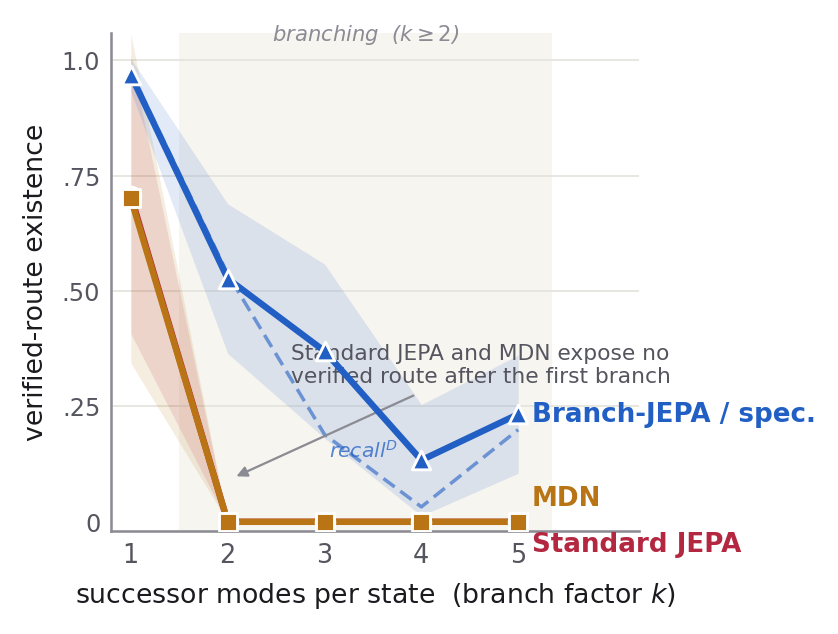}\\[-1mm]
\textbf{\small(c) Branch-factor stress test}
\end{minipage}
\caption{\textbf{OGBench finite-support cases and specialization diagnostics.}
(a) All four immutable archive-B contexts.  (b) At a three-successor state, three
active atoms align with distinct successor latents; five-seed coverage is
$76.1{\pm}23.9\%$, versus $33.3{\pm}0.1\%/32.8{\pm}0.5\%$ for Standard/fused
readouts.  (c) Held-out verified-route existence under increasing branch factor;
$k{=}4/5$ remain unresolved at $13.4{\pm}13.5/23.3{\pm}14.3\%$.}
\label{fig:ogbench-all-cases-supp}
\label{fig:restored-specialization-supp}
\end{figure*}

\begin{figure*}[!t]
\centering
\includegraphics[width=.63\textwidth]{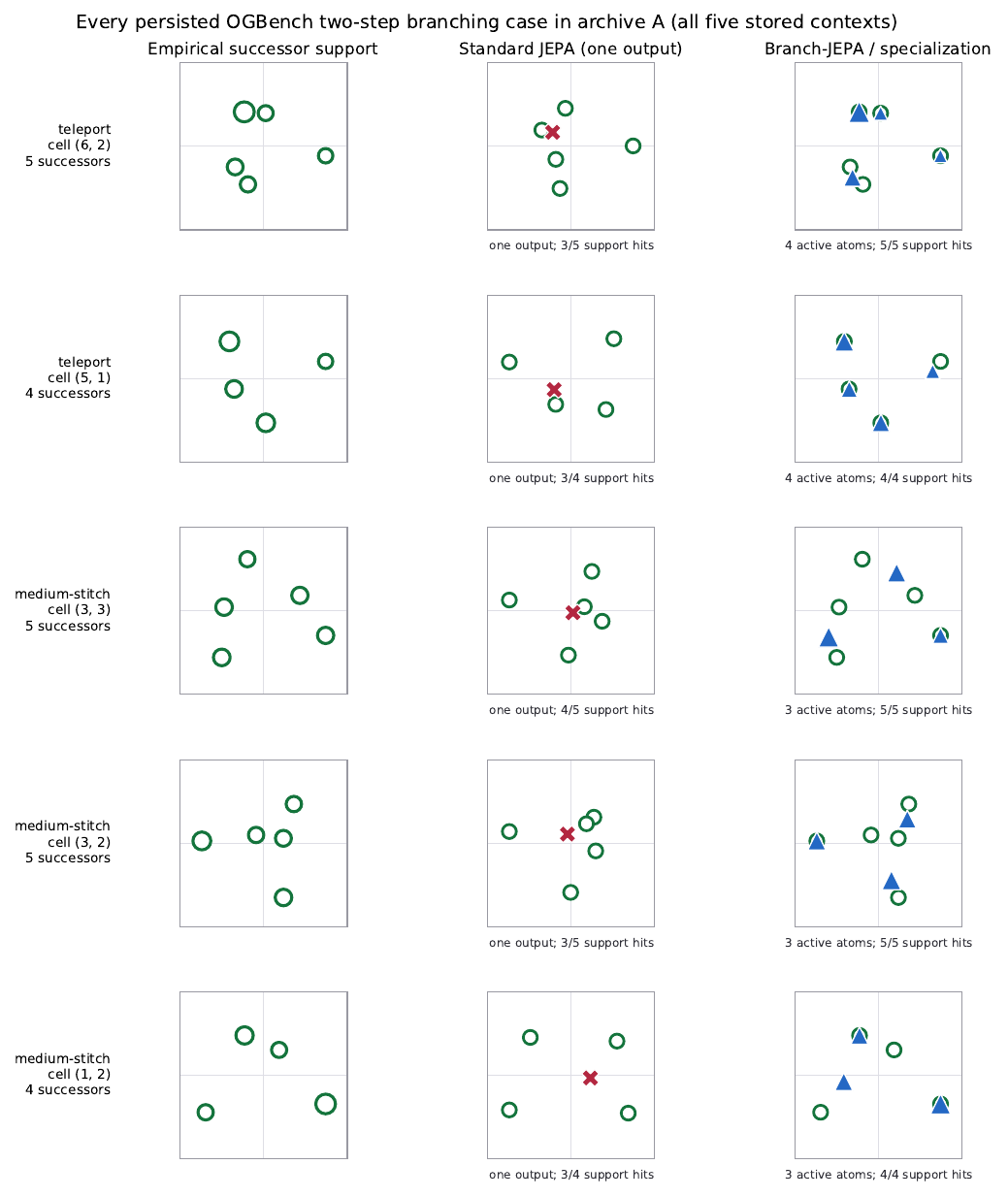}\\[-1mm]
\includegraphics[width=.65\textwidth]{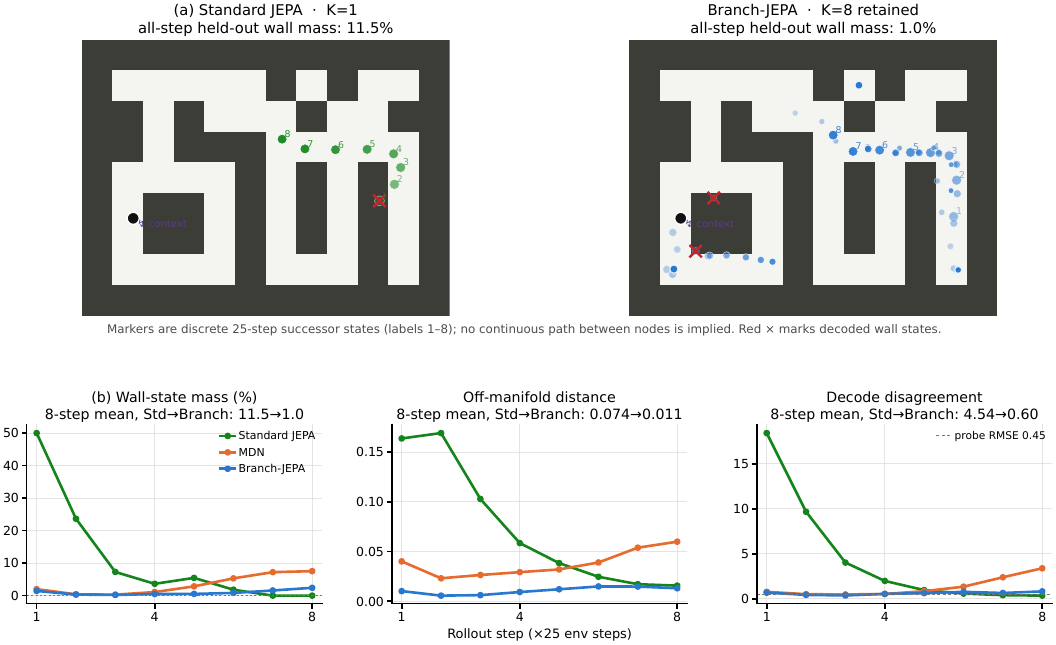}
\caption{\textbf{Additional target-blind support audits.}
(a) All five fixed archive-A contexts: empirical successors, Standard JEPA's single
prediction, and every router-active Branch-JEPA atom.  Empirical support is added only
after emission for this post-hoc projection.  (b) The first frozen held-out AntMaze
context and aggregate curves over all 110 eligible contexts.  Red crosses mark
probe-decoded wall cells; nodes are 25 environment steps apart and do not imply a
continuous collision-free path.}
\label{fig:ogbench-archive5-supp}
\label{fig:antmaze-heldout-rollout-supp}
\end{figure*}

\paragraph{Case-audit protocols.}
The point-maze tensor archives record emissions and context-only router masses before any
realized successor is consulted.  Archive A is an independently trained one-seed extraction
whose five fixed contexts exactly match the checked-in producer's cell list; Figure~\ref{fig:ogbench-archive5-supp}
\textup{(a)} shows all five rather than selecting favorable rows.  Archive B, used in
Figure~\ref{fig:ogbench-all-cases-supp}(a), contains all four contexts from a second extraction.
The two archives are mechanism snapshots, not repeated seeds for a common aggregate.

For the episode-held-out AntMaze audit, a separate one-seed campaign splits 1,000 complete
teleport episodes into 700 training and 300 held-out episodes using a fixed split seed.
At stride 25 and horizon 8, all 110 eligible evaluation contexts come only from held-out
episodes.  Figure~\ref{fig:antmaze-heldout-rollout-supp}\textup{(b)} uses context index zero in the frozen
deterministic ordering, not a context selected using future outcome, model error, or method
margin.  Metrics aggregate every retained branch with its context-only router mass.  This
campaign is separate from the three-seed raw-COV-R campaign in
Table~\ref{tab:interface-benchmarks-supp}.

\paragraph{How to read the OGBench atlases.}
Across all nine archived entries, green circles are empirical successors added after
prediction, red crosses are Standard-JEPA outputs, and blue triangles are all
router-active Branch-JEPA atoms; marker area encodes empirical frequency or context-only
router mass, and a support hit uses the fixed $.5$ cosine threshold.

Figure inputs (SHA-256 prefixes) are \texttt{head\_specialization.json}
(\texttt{4ec6adc3}) and \texttt{branching\_complexity.json} (\texttt{d4b7322f}).

A separate three-seed execution control is inconclusive: medium-stitch replanning is
$86.7{\pm}11.5/66.7{\pm}11.5\%$ for MDN/Branch-JEPA
(mean${\pm}$sample SD); large-stitch and a one-seed DINO-WM pilot are also inconclusive.

\end{document}